\documentclass{article}

\usepackage{microtype}
\usepackage{graphicx}
\usepackage{subcaption}
\usepackage{booktabs} 

\usepackage{hyperref}

\usepackage[preprint]{icml2026}

\usepackage{amsmath}
\usepackage{amssymb}
\usepackage{mathtools}
\usepackage{amsthm}

\usepackage{url}
\usepackage{booktabs}       
\usepackage{amsfonts}       
\usepackage{nicefrac}       
\usepackage{microtype}      
\usepackage{xcolor}         
\usepackage{pifont} 
\usepackage{placeins}
\usepackage{float}
\usepackage{wrapfig}
\usepackage{graphicx}
\usepackage{multicol}
\usepackage{multirow}
\newcommand{\xhdr}[1]{\vspace{2mm}\noindent{{\bf #1.}}}

\usepackage[capitalize,noabbrev]{cleveref}

\theoremstyle{plain}

\theoremstyle{definition}

\theoremstyle{remark}

\usepackage[textsize=tiny]{todonotes}

\icmltitlerunning{Understanding the Dynamics of Demonstration Conflict in In-Context Learning}

\begin{document}

\twocolumn[
  \icmltitle{Understanding the Dynamics of Demonstration Conflict in In-Context Learning}



  \icmlsetsymbol{equal}{*}

  \begin{icmlauthorlist}
  \icmlauthor{Difan Jiao}{tor}
  \icmlauthor{Di Wang}{kaust}
  \icmlauthor{Lijie Hu}{mbzuai}
  \end{icmlauthorlist}

\icmlaffiliation{tor}{University of Toronto}
\icmlaffiliation{kaust}{King Abdullah University of Science and Technology}
\icmlaffiliation{mbzuai}{Mohamed bin Zayed University of Artificial Intelligence}

\icmlcorrespondingauthor{Difan Jiao}{difanjiao@cs.toronto.edu}
\icmlcorrespondingauthor{Lijie Hu}{lijie.hu@mbzuai.ac.ae}





  \vskip 0.3in
]



\printAffiliationsAndNotice{}  

\begin{abstract}

In-context learning enables large language models to perform novel tasks through few-shot demonstrations. However, demonstrations per se can naturally contain noise and conflicting examples, making this capability vulnerable. To understand how models process such conflicts, we study demonstration-dependent tasks requiring models to infer underlying patterns, a process we characterize as rule inference. We find that models suffer substantial performance degradation from a single demonstration with corrupted rule. This systematic misleading behavior motivates our investigation of how models process conflicting evidence internally. Using linear probes and logit lens analysis, we discover that under corruption models encode both correct and incorrect rules in intermediate layers but develop prediction confidence only in late layers, revealing a two-phase computational structure. We then identify attention heads for each phase underlying the reasoning failures: Vulnerability Heads in early-to-middle layers exhibit positional attention bias with high sensitivity to corruption, while Susceptible Heads in late layers significantly reduce support for correct predictions when exposed to the corrupted evidence. Targeted ablation validates our findings, with masking a small number of identified heads improving performance by over 10\%. 




\end{abstract}
\section{Introduction}

In-context learning (ICL) enables large language models (LLMs) to adapt to new tasks through few-shot demonstrations without parameter updates \citep{brown2020language}. A particularly intriguing aspect of ICL is the capability to infer underlying patterns from demonstrations, marking a form of general intelligence akin to human reasoning \citep{dong2022survey, bai2023transformers, kossen2023context}. For example, LLMs are able to learn and perform competitively on linear regression and analogical reasoning problems, demonstrating their capacity to learn abstract patterns from limited contextual information \citep{von2023transformers, coda2023meta, zhang2024trained, fu2025short}.

\begin{figure*}[t]
    \centering
    \includegraphics[width=0.99\linewidth]{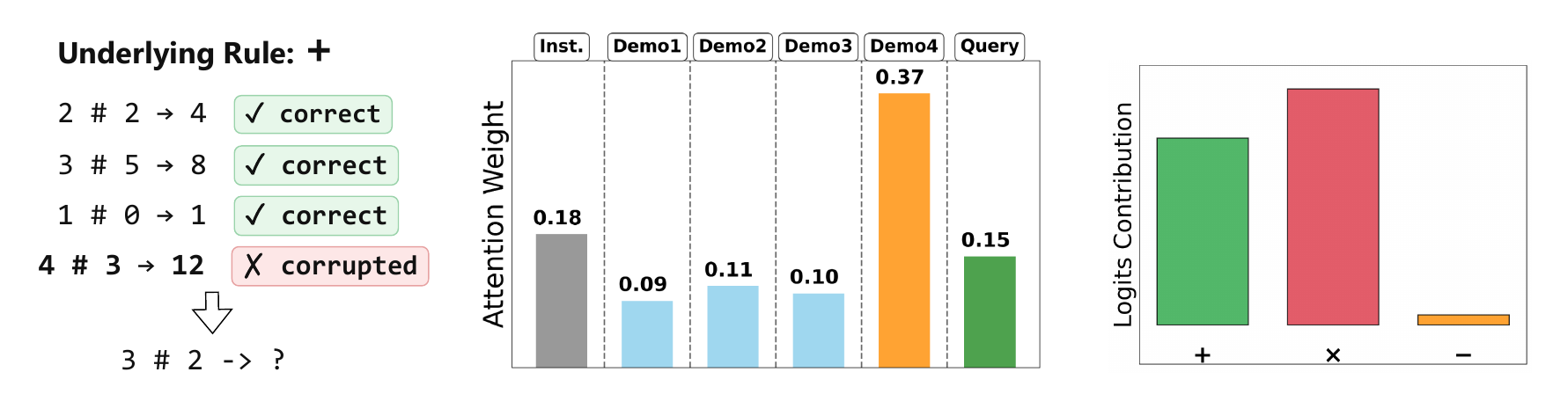}
    \vspace{-0.2cm}
    \caption{\textbf{Left:} The Operator Induction task where models are required to infer the underlying rule (+) from demonstrations, among which we introduce one corrupted example.  \textbf{Middle:} Vulnerability heads show disproportionate attention allocation; they also show drastic output changes when the heavily-attended position corrupts. \textbf{Right:} Susceptible heads shows logit contributions favoring the corrupted operator ($\times$) over the correct one (+) despite minority corruption.}
    \label{fig:teaser}
    \vspace{-0.4cm}
\end{figure*}

However, learning the novel rules in-context can be intrinsically vulnerable to conflicting demonstrations, with ICL performance broadly known to degrade substantially when demonstrations contain contradictory information \citep{yoo2022ground, min2022rethinking, wei2023larger}. This vulnerability is particularly concerning given that real-world data naturally contains noise and outliers that can mislead models' judgment about underlying patterns. While previous work has examined the conflict resolution in general tasks such as text classification and question answering \citep{li2023contradoc, xie2023adaptive, halawi2023overthinking, yu2024large}, these findings cannot be directly translated to rule inference settings. Notably, unlike the tasks where models already possess sufficient parametric knowledge \footnote{known as the factual and relational information that an LLM acquires after pre-training.} for reliable zero-shot performance \citep{yang2022empirical, jiao2023spin}, rule inference requires genuine demonstration reliance.

In this work, we first investigate the dynamics of in-context rule inference under conflicting demonstrations. We employ two tasks: Operator Induction \citep{zong2024vl} and Fake Word Inference (mapping synthetic vocabulary to real concepts), where models perform both tasks well with ICL but can only achieve chance-level performance without in-context examples. In these tasks, the modularity of demonstrations allows us to introduce conflicts by corrupting rules at specific positions without undermining others. Our experiments reveal that LLMs suffer substantial performance degradation from a single corrupted example among a vast majority of correct ones. We also observe systematic performance variance when corrupting different positions of demonstrations, also known as the positional bias \citep{wang2024eliminating, cobbina2025show}

During evaluation, we notice a strong signal that models are misled by the injected conflict. When corruption flips models from correct to wrong, the corrupted operator notably accounts for predictions despite appearing in only one position. Given this, we speculate that models encode both rules during the process, yet can fail to resolve the correct rule at the end. This drives us to study how conflicting evidence evolves and affects model behaviour during the internal processing. First, we train linear probes \citep{alain2016understanding} to track the emergence of each operator in internal representations, showing that the models encode both correct and misleading rule in intermediate layers. Then, we run logit lens \citep{nostalgebraist2020logitlens} to decode model predictions layer by layer, revealing that models develop strong confidence for both rules at the final layers. This temporal separation suggests a natural two-phase hypothesis: conflict creation and conflict resolution.

To test this hypothesis, we identify model components responsible for each phase. For the first phase, since demonstrations are independent and equivalent in our tasks, the positional bias indicates that specific components treat positions unequally. Building on prior work that attributes this bias to attention heads \citep{wang2024eliminating}, we localize \textbf{Vulnerability Heads} (Figure \ref{fig:teaser} middle) that both attend disproportionately to certain positions and show high sensitivity when those positions are corrupted, finding their concentration in early-to-middle layers. For the second phase, we localize \textbf{Susceptible Heads} (Figure \ref{fig:teaser} right) whose preference for the correct over corrupted operator decreases the most when the corruption is introduced, finding their concentration in later layers. Ablating these two types of heads mitigates the performance degradation to more than 10\%, confirming the causal roles of both head types in creating and falsely resolving conflicts. Finally, our correlational test reveals a synergy between two heads: masking vulnerability heads significantly reduces the contribution of susceptible heads towards the corrupted operator.


Our contributions are twofold. First, we establish a foundational framework for studying the dynamics of conflict resolution during in-context rule inference. Second, we provide mechanistic evidence for a two-phase structure of reasoning under corruption, identifying specific model components that exhibit both correlational and causal relevance to conflict creation and conflict resolution.

\section{Related Work}

\subsection{Conflict Resolution in In-Context Learning}

Our work studies scenarios where LLMs encounter conflicts during in-context learning. Research indicates two types of conflicts emerging in ICL \citep{xu2024knowledge}: context-memory conflicts where contextual information contradicts the model's parametric knowledge, and inter-context conflicts where multiple demonstrations contradict. While both conflicts are common in real world \citep{xu2024knowledge}, prior investigations predominantly focus on how context-memory conflicts lead to wrong predictions in tasks including text classification, textual and visual question answering, etc. \citep{pan2021attacking, min2022rethinking, xie2023adaptive, halawi2023overthinking, hua2025vision,zhang2025modalities,zhang2024locate,cheng2024multi}. Beyond understanding these conflicts, several work has explored mitigation strategies through adversarial robustness analysis \citep{wang2023adversarial} and artificial demonstration selections \citep{li2024debiasing, qin2024context, hu2024strategic}. In contrast, we focus on inter-context conflicts in in-context rule-learning scenarios. While models demonstrate strong capabilities on rule inference tasks like operator induction, linear regression, and analogical reasoning \citep{von2023transformers, coda2023meta, zong2024vl}, how conflicting demonstrations affect their reasoning dynamics remains underexplored.

\subsection{Mechanistic Interpretability of In-Context Learning}

Our approach to understanding conflict processing mechanisms follows the work of mechanistic interpretability, which seeks to reverse-engineer the computational framework underlying neural network behaviors \citep{olah2020zoom, elhage2021framework, cammarata2021curve,zhang2025eap}. These methods enable us to identify when and where models encode specific information, trace how representations evolve across layers, and localize model components responsible for particular behaviors \citep{yang2024exploring,zhang2025understanding,meng2022locating, jiao2023spin, hu2024hopfieldian,su2025understanding,dong2025understanding}. 

Recent work applying mechanistic interpretability to ICL has revealed foundational insights about demonstration processing. \citet{wang2023label} analyze ICL through an information flow lens, identifying label words as anchors where semantic information aggregates in shallow layers before serving for predictions, and \citet{liu2023towards} employs saliency maps to analyze the importance of sentence components during ICL. Notably, a thread of studies focus on functional attention heads for interpreting ICL, following the induction heads \citet{olsson2022context}, which are crucial for pattern completion and copying behaviors in ICL. Studies have identified specific attention heads responsible for various ICL phenomena: mechanisms underlying few-shot learning capabilities \citep{wang2022interpretability}, heads that promote incorrect predictions under adversarial contexts \citep{halawi2023overthinking}, and components that contribute to positional bias \citep{han2024unibias}. This body of work establishes attention heads as primary computational substrates for ICL processing, providing the foundation for our component-level analysis of conflict creation and conflict resolution.

\section{Dynamics in ICL under Conflicting Demonstrations}

\begin{figure*}[t!]
    \centering
    \includegraphics[width=0.99\linewidth]{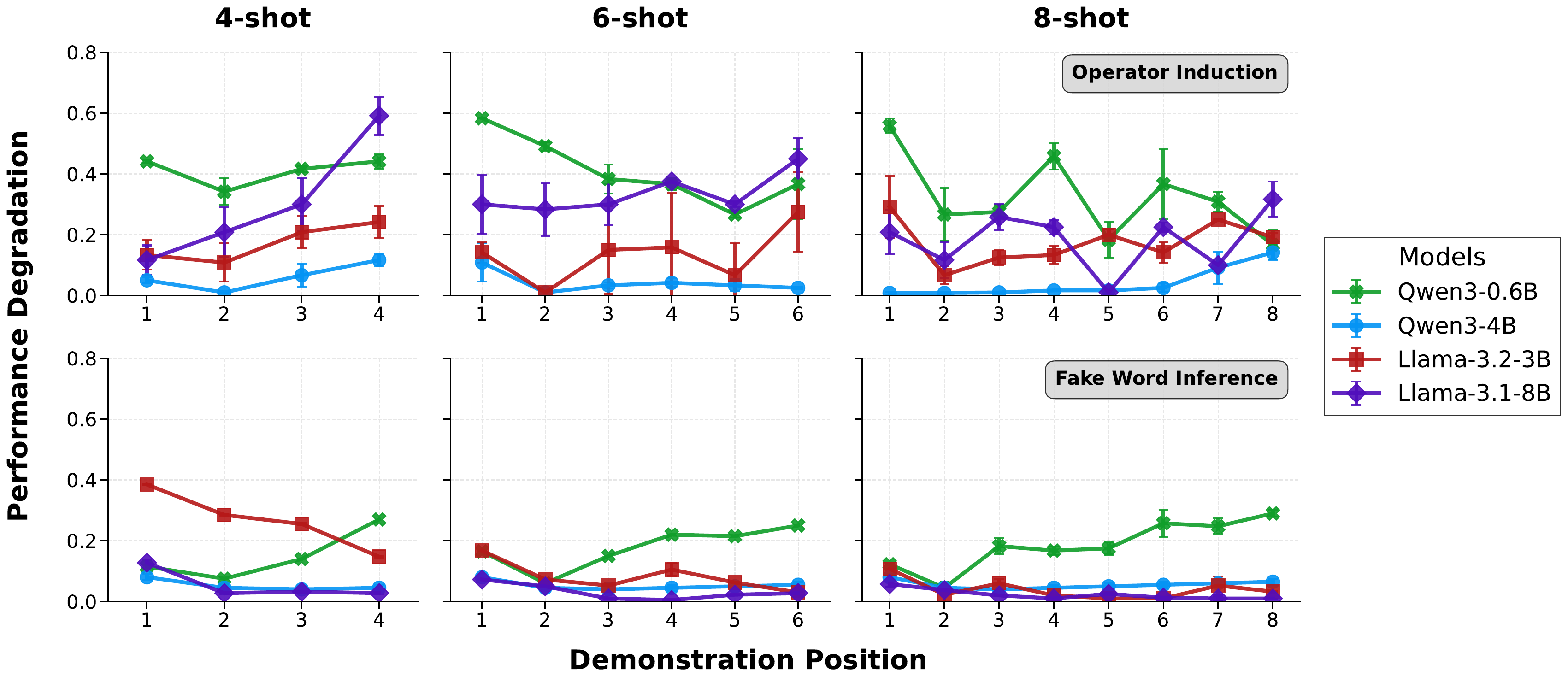}
    \vspace{-0.2cm}
    \caption{Performance degradation under single-position corruption across different large language models and tasks. Each point represents the decrease in accuracy when corrupting the demonstration at that specific position.}
    \label{fig:performance_degradation_main}
    \vspace{-0.6cm}
\end{figure*}

We start by formulating the task of in-context rule inference that we study: LLMs relying on few-shot demonstrations to infer the underlying rule of a novel task. To observe model's reasoning dynamics under conflict, we establish an intervention framework that introduces targeted conflicts among demonstrations. 

\subsection{A Corruption-Based Intervention Framework}

\xhdr{Task Setup} Without loss of generality, we formulate the task as where language models are instructed to infer latent rules from demonstrations $\mathcal{D} = \{(x_i, y_i)\}_{i=1}^k$ and apply the learned rule to predict $\hat{y}_q = r^*(x_q)$ for query $x_q$. Clean demonstrations represent unanimous evidence for the underlying rule $r^* \in \mathcal{R}$, where $\mathcal{R}$ is the hypothesis space of possible rules. To introduce conflicting evidence among demonstrations, we intervene with a position-specific corruption at demonstration index $p \in \{1, \ldots, k\}$, by replacing the correct output $y_p = r^*(x_p)$ with $y_p^{\text{corrupt}} = r'(x_p)$ where $r' \neq r^*$. The model is then required to do the exact same task under the corrupted demonstration set as:

\vspace{-0.9cm}
\begin{equation}
\begin{split}
\mathcal{D}_p^{\text{corrupt}} = \{ & (x_1, y_1), \ldots, (x_p, y_p^{\text{corrupt}}), \\
& \ldots, (x_k, y_k) \}
\end{split}
\end{equation}
\vspace{-0.5cm}

This setup carries three fundamental principles. First, we require \textbf{demonstration reliance}, where models are expected to exhibit chance-level performance in zero-shot settings, ensuring their genuine reliance on demonstrations rather than on their parametric knowledge. Secondly, during the evaluation of performance under corruptions, we impose a \textbf{majority rule} by introducing conflicting demonstrations only when correct ones substantially outnumbers corrupted ones. Thirdly, we require the \textbf{modularity} of demonstrations during inference, where each demonstration is independent and provides equally weighted evidence of the underlying rule. Collectively, these three principles provide a valid foundation for studying reasoning dynamics through which LLMs inference rules with contradictions.


\xhdr{Datasets, Models, and Evaluation} Note that a wide variety of ICL tasks either involve interdependent demonstrations or potentially provide unequal weights per demonstration \citep{wei2021finetuned}, we carefully adopt two tasks that adhere to our principled design criteria: Operator Induction~\citep{zong2024vl} where models must infer the underlying mathematical operation from demonstrations to answer the final query, and Fake Word Inference where models learn mappings from synthetic vocabulary to real concepts (e.g., "blimontar glemorax" $\rightarrow$ "red hat"). We conduct experiments across four LLMs, including Qwen3 models (Qwen3-0.6B, Qwen3-4B) \citep{yang2025qwen3} and Llama3 models (Llama-3.2-3B-Instruct, Llama-3.1-8B-Instruct) \citep{dubey2024llama}. We interpret model's performance gap between clean and corrupted runs as reflecting the difficulty in autonomously detecting and resolving conflicts. Details of datasets and the evaluation pipeline are included in Appendix \ref{app:framework}.


\subsection{Consistent Performance Degradation with Single-Position Corruption}

Following our corruption framework, we systematically evaluate model performance with different numbers of demonstrations across all tasks. We first validate models' in-context learning capability for the tasks with no corruptions. Then, to observe performance dynamics under conflicts, we run position-specific corruption by corrupting only a single demonstration at each position within the demonstration sequence and examine the calibrated performance gap position by position, maintaining the majority rule. As shown in Table \ref{tab:baseline_performance} at Appendix \ref{app:framework}, all models demonstrate reliable in-context performance with clean demonstrations. However, as shown in Figure \ref{fig:performance_degradation_main}, LLMs exhibit consistent and substantial performance degradation under conflicts. The degradation can reach as high as 58 percentage points for Operator Induction, with an average degradation of 16 percentage points across all cases. This degradation persists across model scales and architectures, and notably, performance degradation persists even as the number of demonstrations increases to 8, where correct samples maintain a dominant 7:1 advantage over conflicting ones. This pattern indicates that models are significantly obfuscated by conflicting, noisy information during in-context rule inference.



\xhdr{Models Heavily Adopt Corrupted Rules} To understand how models fail under corruption, we analyze the pattern of wrong predictions when models flip from correct to incorrect answers. By examining which rule underlies each wrong prediction, we can determine whether models are systematically misled by corrupted evidence or failing randomly. Importantly, we observe that the corrupted operator accounts heavily for these wrong answers, notably 81\% from Qwen3-4B and 71\% from Llama-3.2-3B in Operator Induction, despite appearing in only one demonstration position. This systematic adoption of minority corrupted evidence, rather than arbitrary errors, suggests that models actively encode and process conflicting rules during inference, yet fail to properly resolve which rule to apply for final predictions.

\xhdr{Positional Bias Emerges from Corruption Patterns} Moreover, our experimental design reveals notable positional bias \citep{hsieh2024found, cobbina2025show} as shown in Figure \ref{fig:performance_degradation_main}, where identical corruptions produce different performance impacts depending on their location within the demonstration sequence. This observed bias is particularly instructive in our framework, given the independence and equivalence nature of demonstrations. Under ideal conflict resolution, models should exhibit uniform vulnerability to corruption across all positions, yielding near-zero position bias. The authentic nature of this position bias, uncoupled from confounds of demonstration-specific and parametric knowledge reliance, provides a clean signal for our analysis.

\section{Internal Mechanisms of ICL Under Conflicting Demonstrations}

In the previous section, we present evaluation experiments that reveal clear LLM performance degradation under corruption. However, these evaluations provide little insight into when and how LLMs internally represent and resolve conflicting demonstrations during in-context learning. In this section, we aim to understand these questions by analyzing model internal representations. 

\subsection{LLMs Encode Underlying Rules in Intermediate Layers}

We use linear probes~\citep{alain2016understanding} to identify where models encode the rule information. To investigate the evolution of rule encoding, we train linear probes to identify each possible underlying rule in the residual stream across all transformer layers. Details of the probe training setup and validation can be found in Appendix \ref{app:linear_probes}.

\begin{figure}
\vspace{-0.0cm}
\centering
\includegraphics[width=0.95\linewidth]{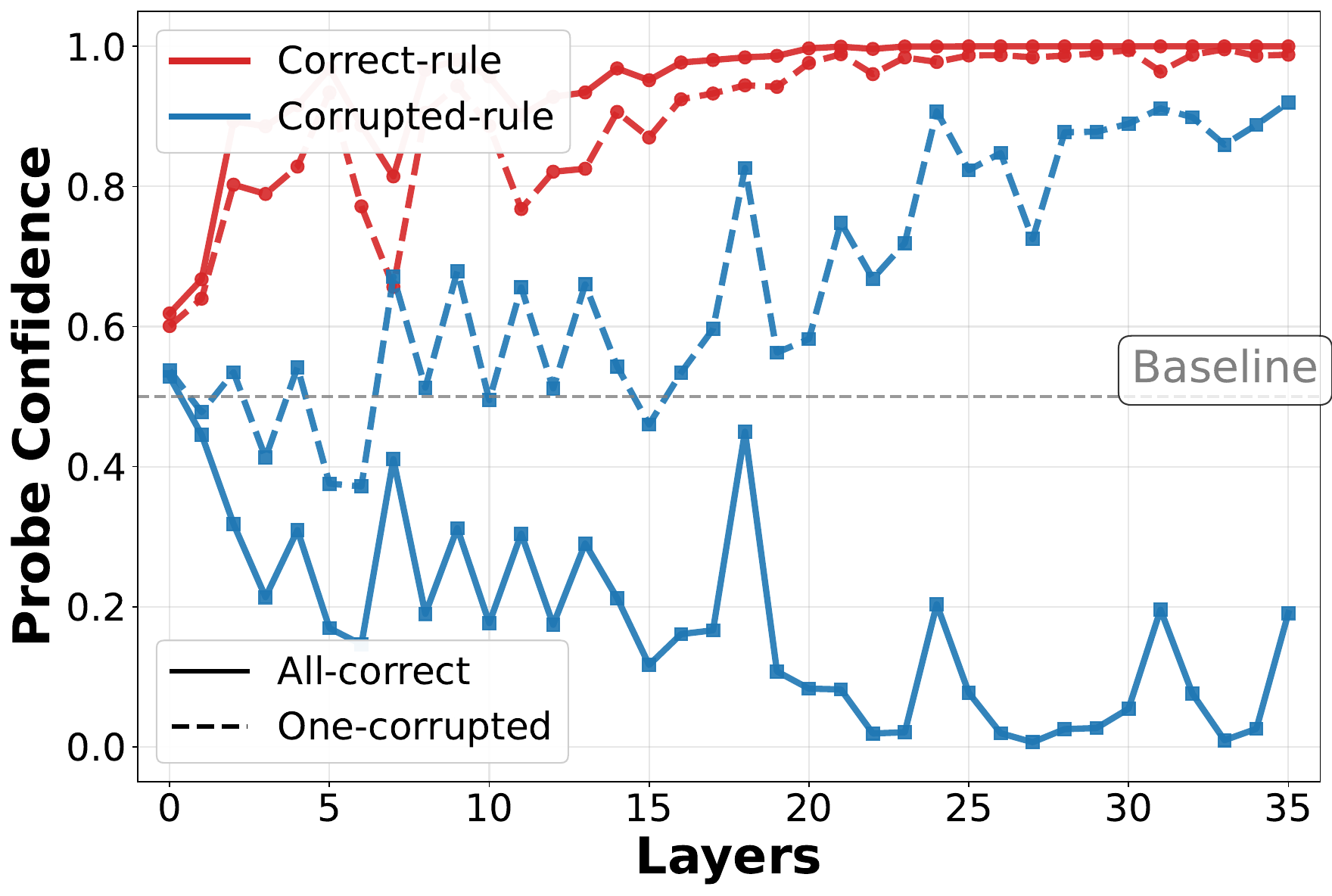}
\caption{Linear probe confidence across model layers under different corruption scenarios. The Correct Probe detects model's encoding of ground-truth operator while the Corrupted Probe detects the corrupted operators.}
\label{fig:probe_analysis_main}
\vspace{-0.8cm}
\end{figure}

We then evaluate model's encoded confidence of detecting the rules from demonstrations, measured by the predictions of linear probes, under two corruption scenarios: no corruption (all demonstrations are correct) and minority corruption (correct demonstrations outnumber corrupted ones). As shown in Figure \ref{fig:probe_analysis_main}, probing results reveal two patterns. First, when no corruption information is demonstrated at all, the model shows consistent evidence encoding with high confidence for the correct rule and low confidence for corrupted rules across layers. On the other hand, as corrupted demonstrations are introduced, probe confidence for corrupted rules rises significantly above baseline, indicating that models actively identify and encode the corrupting rule in the process. Notably, internal representations of both correct and incorrect rules emerge most prominently in early-to-middle layers, with confidence levels becoming more stable in later layers. This suggests that models encode multiple competing rules simultaneously during the intermediate processing stages, potentially creating an internal representational conflict that needs to be resolved to produce coherent final predictions.

\subsection{LLMs Resolve Conflicting Evidence in Late Layers}

While probes reveal where models encode information, they cannot determine when models actively use the information for predictions \citep{elazar2021amnesic}. To observe the dynamics of model's favoring specific rules during ICL inference, we employ logit lens by projecting internal representations through the unembedding matrix to decode model predictions at intermediate layers \citep{nostalgebraist2020logitlens}. In more detail, to make the unembedded predictions more interpretable, we modify our ICL task prompts to elicit LLMs to identify the underlying rule rather than the final result.

\begin{figure}
\vspace{-0.0cm}
\centering
\includegraphics[width=0.98\linewidth]{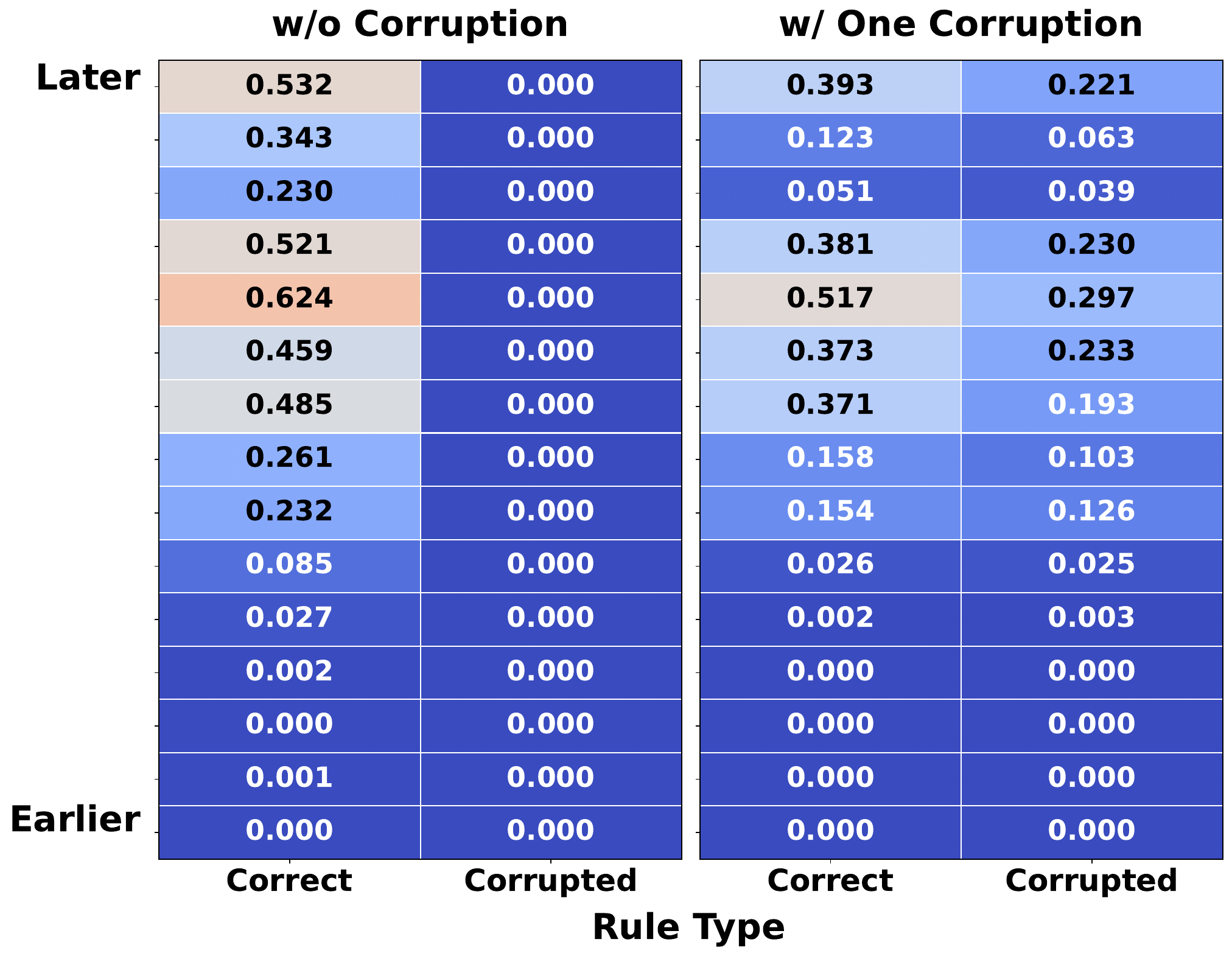}
\caption{Logit lens predictions across corruption scenarios. Rows show different layers, columns represent correct and corrupted rules, and values indicate prediction probability for each rule decoded from layer-wise residual streams.}
\label{fig:logit_lens_main}
\vspace{-0.6cm}
\end{figure}

For consistency, we collect logit lens outputs under the same two corruption scenarios as linear probes. First, we notice a sharp transition in prediction probability occurring at late layers, where the model suddenly develops strong confidence for specific rules. In contrast, we observe that the model empirically exhibits minuscule prediction confidence across both correct and corrupted rules in early layers, and thus we omit those layers in Figure \ref{fig:logit_lens_main}. Secondly, similar to probing results, we observe the simultaneous emergence of confidence for both correct and corrupted rules under the only-one-corruption setup, further indicating an existing competition between conflicting rules.

\subsection{A Two-Stage Hypothesis of ICL Reasoning Dynamics under Corruption}

Synthesizing the linear probe and logit lens findings, we observe a temporal separation between rule encoding (early-middle layers) and prediction formation (late layers). This temporal structure prompts a hypothesis that model's reasoning under in-context corruptions involves two complementary computational phases, which we term \textit{conflict creation} and \textit{conflict resolution}.

We characterize this two-stage hypothesis in more detail. In the first phase, certain model components create systematic weak points by encoding corrupted information, making the system vulnerable to conflicts. In the second phase, the conflict resolution process itself can fail when components are overly susceptible to minority corrupted evidence, leading to incorrect final predictions despite majority support for the correct rule. This framework predicts that early-layer heads should exhibit high sensitivity to positional corruption (creating vulnerabilities), while late-layer heads should demonstrate susceptibility patterns that undermine robust majority-based decisions (resolution failures). We test this hypothesis by identifying components associated with each failure mode and examining their causal contributions to reasoning under corruption.

\subsubsection{Locating Conflict Creation}
\label{understanding:creation}

\begin{figure*}[t!]
    \centering
    \includegraphics[width=0.90\linewidth]{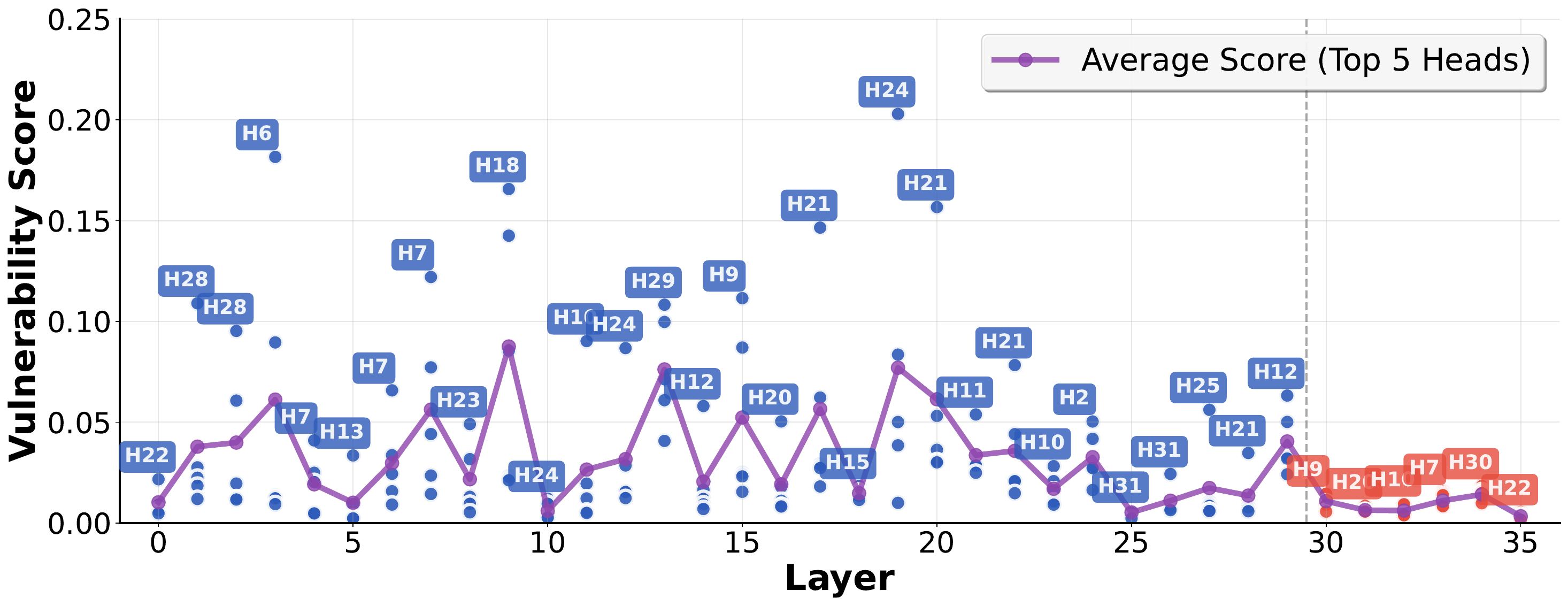}
    \vspace{-0.0cm}
    \caption{Distribution of Vulnerability Heads across model layers. The purple line shows the average vulnerability score (A × S) among the top 5 heads per layer. Individual heads are shown as dots.}
    \label{fig:vulnerability_heads_distribution}
    \vspace{-0.5cm}
\end{figure*}

We begin by analyzing the first phase of our hypothesis: how certain model components create systematic vulnerabilities under corruption. Under ideal reasoning, each demonstration should contribute equally to rule inference due to the demonstration-dependent nature and equivalent complexity across demonstrations, yielding uniform positional effects. 

However, we observe systematic positional variances in performance degradation as shown in Figure \ref{fig:performance_degradation_main}. This positional bias is not new to research, with prior studies attributing it mostly to the disproportionate integration from demonstrations through the attention mechanism \citep{han2024unibias, peysakhovich2023attention, chen2023positional}. Building on prior work, we characterize two measurable attributes of attention heads that potentially contribute to the position-variant degradation under corruption: (1) disproportionate attention allocation per position, and (2) high output sensitivity when heavily-attended positions are corrupted. In other words, these heads are prone to create systematic vulnerabilities when contradictory information appears at their heavily-attended positions. To quantify them, we follow the ICL prompt segmentation from \citet{cho2024revisiting}, and focus our analysis on the final query forerunner token, which is identified critical for encoding input text representations during in-context learning.

\xhdr{Measuring Positional Attention Allocation} For each attention head $(l,h)$ at layer $l$ and head $h$, we extract the attention score $\mathbf{A}_{l,h} \in \mathbb{R}^{S \times S}$ where $S$ represents sequence length. Following the demonstration-query segmentation \citep{cho2024revisiting}, we compute the attention allocation from the query forerunner token position $\mathbf{F}$ to each demonstration position $p \in \{1, 2, \ldots, k\}$:

\vspace{-0.4cm}
\begin{equation}
    A_{l,h}^{(p)} = \frac{1}{|D_p|} \sum_{t \in D_p} \mathbf{A}_{l,h}[\mathbf{F}, t]
\end{equation}
\vspace{-0.4cm}

where $D_p$ denotes the set of token positions belonging to demonstration $p$, and $|D_p|$ represents the number of tokens in that demonstration. This averaging yields per-demonstration attention scores that capture how much each head attends to specific positions during query processing.

\begin{figure}
\centering
\vspace{-0.3cm}
\includegraphics[width=0.95\linewidth]{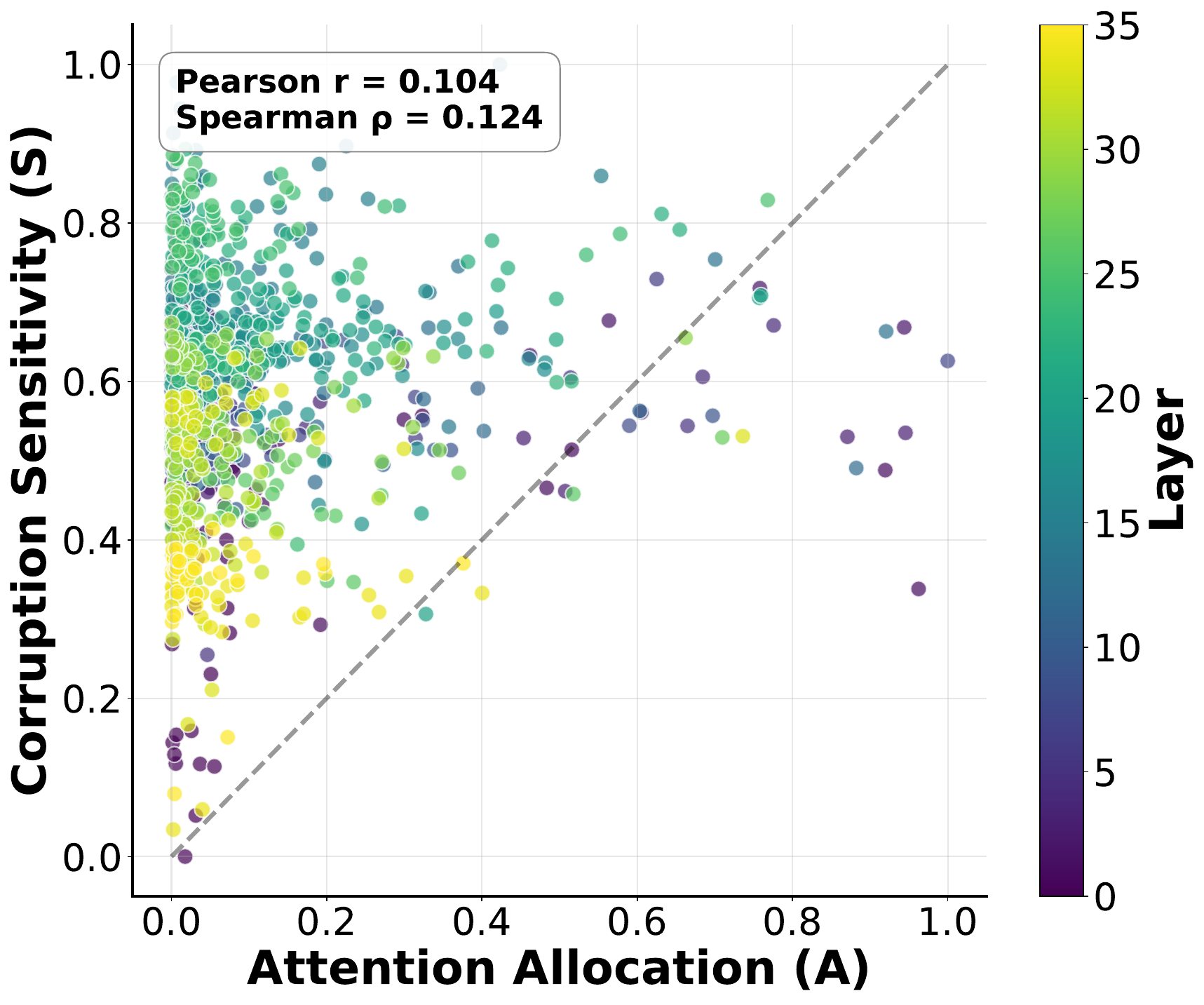}
\caption{Weak correlation revealed between attention score allocation (A) and corruption sensitivity (S).}
\label{fig:two_metrics_corr}
\vspace{-0.5cm}
\end{figure}

\xhdr{Measuring Attention Head Sensitivity to Corruption} For each attention head $(l,h)$ and demonstration position $p$, we compute sensitivity as the normalized change in head outputs when that position is corrupted. Specifically, we first extract the head's output $\mathbf{o}_{l,h}^{(\mathbf{F})}(\text{clean})$ at the query forerunner position $\mathbf{F}$ under all-correct conditions, then corrupt only position $p$ with incorrect rule variants $r \in \mathcal{R}_{\text{wrong}}$ while keeping all other positions unchanged. We measure the output deviation $\mathbf{o}_{l,h}^{(\mathbf{F})}(\text{corrupted})$ for each corruption and compute the average normalized change. The sensitivity score is formulated as:

\begin{figure*}[t!]
    \centering
    \includegraphics[width=0.90\linewidth]{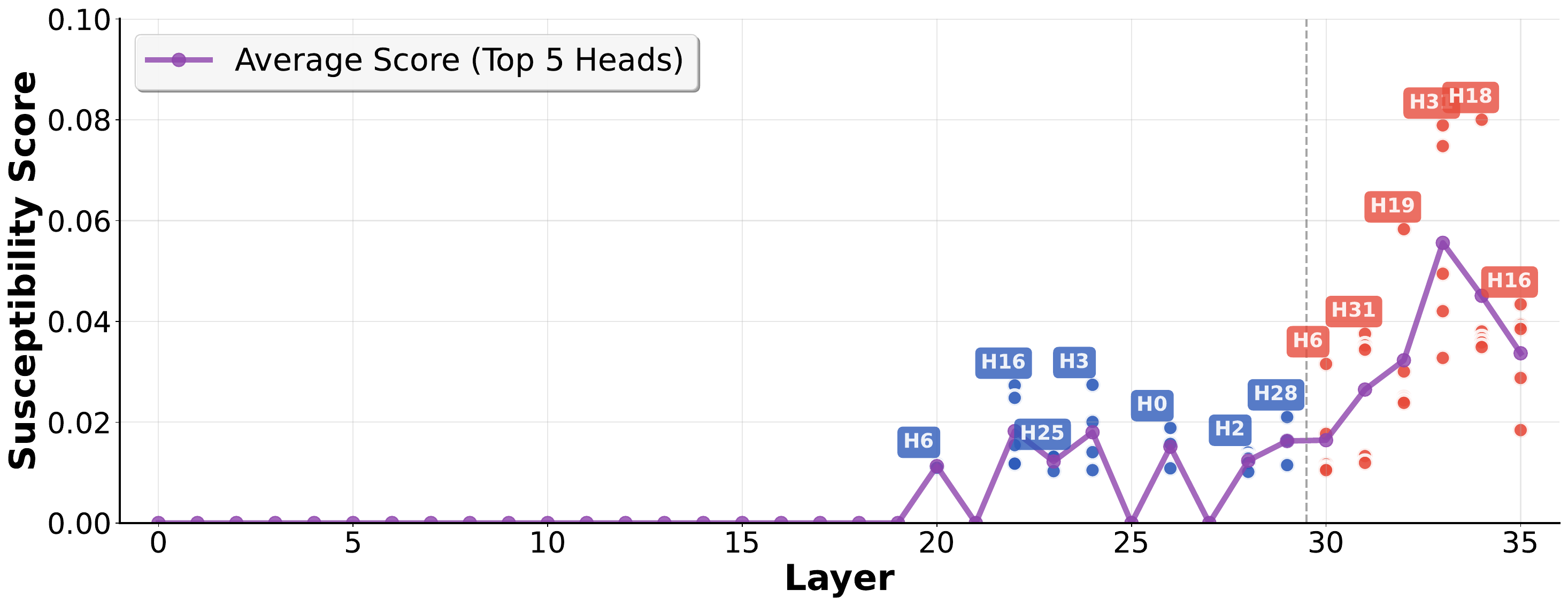}
    \vspace{-0.0cm}
    \caption{Distribution of susceptible heads across model layers. The purple line shows average susceptibility scores among top 5 heads per layer. Individual heads shown as dots.}
    \label{fig:conflict_resolution_distribution}
    \vspace{-0.5cm}
\end{figure*}

\vspace{-0.25cm}
\begin{equation}
    S_{l,h}^{(p)} = \frac{\left\|\mathbf{o}_{l,h}^{(\mathbf{F})}(\text{corrupt}) - \mathbf{o}_{l,h}^{(\mathbf{F})}(\text{clean})\right\|_2}{\left\|\mathbf{o}_{l,h}^{(\mathbf{F})}(\text{clean})\right\|_2 + \epsilon}
\end{equation}
\vspace{-0.25cm}

High sensitivity indicates heads whose outputs fluctuate substantially with conflicting demonstrations. Intuitively, for corrupted position $p$, we are interested in localizing those attention heads that attend heavily to $p$ while being extremely sensitive in outputs to the single corruption at $p$. Before combining these attributes, we first assess whether they measure the same or distinct facets of attention head behaviour. To this end, we quantify their dependence using Pearson and Spearman correlations, with results shown in Figure \ref{fig:two_metrics_corr}. We observe weak associations across heads, indicating that our two characterizations are largely complementary rather than redundant. Thus, we multiply them as the \textit{Vulnerability Score} and serve it as a filter for salient heads. Figure \ref{fig:vulnerability_heads_distribution} shows the distribution of Vulnerability Heads for Qwen3-4B, with blue indicating early-to-middle layer heads and red indicating final-layer ones. We observe a clear stratification with vulnerability behavior concentrated in early layers, supporting the first phase of our two-stage framework. However, we point out that this score only reflects correlational patterns rather than causal impacts.

To validate the causal role of these heads in creating vulnerabilities, we conduct targeted ablation experiments by masking identified vulnerability heads during inference. We run three rollouts
for random corruption rules and compare with random heads ablated. As shown in Table \ref{tab:head_ablation_main_OI}, ablating top vulnerability heads improves model performance under corruption, achieving up to 11.12\% relative performance improvement when masking a few vulnerability heads. This substantial downstream effect validates our identification method for vulnerability heads and the contributions of these heads to model's reasoning vulnerabilities under demonstration corruption. 



\subsubsection{Locating Conflict Resolution}

We next investigate the second phase of our hypothesis: how certain model components cause the failure of conflict resolution. Under the corruption framework, robust conflict resolution requires maintaining support for correct predictions even when minority corrupted evidence appears, and we characterize the susceptibility to corruption as the opposite pattern of being easily swayed. Intuitively, and also established by previous studies, attention heads are the primary substrate for mediating how models selectively incorporate information from different demonstrations \citep{olsson2022context, halawi2023overthinking, han2024unibias}. Thus, we focus our analysis on identifying attention heads that exhibit this susceptibility pattern.

\xhdr{Measuring Susceptibility via Logit Attribution} We employ logit attribution \citep{wang2022interpretability} to quantify each attention head's contribution to final predictions under the same two corruption scenarios as above. For each query, we construct two contrastive contexts: clean context where all demonstrations are correct, and minority corruption context where only one demonstration is corrupted. We extract each attention head's output at the query forerunner position and compute its attribution to the token of the correct rule, measured by prediction logits, through the unembedding matrix. Details of logit attribution computation are discussed in Appendix \ref{app:logit_lens}.

Then, we define each head's \textit{Susceptibility Score} as the reduction in its preference for the correct over corrupted rule when exposed to minority corruption. A high positive score indicates that the head is easily swayed by conflicting evidence despite the overwhelming majority supporting the correct rule, suggesting problematic susceptibility rather than robust conflict resolution. We term heads with high susceptibility scores \textit{Susceptible Heads}. Figure \ref{fig:conflict_resolution_distribution} from Qwen3-4B reveals a clear pattern that most salient susceptible heads are predominantly concentrated in late layers. This late-layer concentration contrasts sharply with early-to-middle-layer vulnerability heads, supporting clear functional separation of conflict creation and conflict resolution phases. Since logit attribution does not guarantee causal influence, we further validate these findings through targeted ablation experiments, similar to vulnerability heads. As shown in Table \ref{tab:head_ablation_main_OI}, ablating susceptible heads also improves performance under corruption consistently, achieving up to 10.33\% relative performance improvement when masking a few susceptible heads as well, confirming their causal role in reasoning failures.

\begin{table}[t]
\caption{Attention head ablation results showing relative performance improvement (\%) for Operator Induction, when masking different numbers and different types of attention heads.}
\label{tab:head_ablation_main_OI}
\centering
\addtolength{\tabcolsep}{-3.5pt} 
\begin{tabular}{c c c c c}
\hline
\multirow{2}{*}{\textbf{Models}} & \multirow{2}{*}{\textbf{Heads}} & \multicolumn{3}{c}{\textbf{\# of Ablated Heads}} \\
\cline{3-5}
& & \textbf{5} & \textbf{8} & \textbf{10} \\
\hline
\multirow{3}{*}{\shortstack{Qwen3\\-0.6B}} & Vul. & 1.58{\scriptsize$\pm$1.57} & 4.91{\scriptsize$\pm$1.55} & 8.49{\scriptsize$\pm$1.58} \\ 
& Susc. & 10.33{\scriptsize$\pm$1.48} & 8.41{\scriptsize$\pm$1.18} & 8.21{\scriptsize$\pm$1.50} \\
& Random & -0.85{\scriptsize$\pm$0.42} & -2.34{\scriptsize$\pm$0.38} & -4.67{\scriptsize$\pm$0.51} \\
\hline
\multirow{3}{*}{\shortstack{Qwen3\\-4B}} & Vul. & 1.94{\scriptsize$\pm$0.87} & 0.88{\scriptsize$\pm$0.66} & 1.58{\scriptsize$\pm$0.90} \\ 
& Susc. & 1.58{\scriptsize$\pm$0.90} & 1.94{\scriptsize$\pm$0.87} & 1.94{\scriptsize$\pm$0.87} \\
& Random & -0.44{\scriptsize$\pm$0.20} & -2.07{\scriptsize$\pm$0.85} & -4.24{\scriptsize$\pm$1.03} \\
\hline
\multirow{3}{*}{\shortstack{Llama-3.2\\-3B}} & Vul. & -0.03{\scriptsize$\pm$1.47} & 11.12{\scriptsize$\pm$1.34} & 8.21{\scriptsize$\pm$1.38} \\ 
& Susc. & 0.93{\scriptsize$\pm$1.46} & 4.34{\scriptsize$\pm$1.42} & 1.33{\scriptsize$\pm$1.50} \\
& Random & -0.67{\scriptsize$\pm$0.39}  & -2.15{\scriptsize$\pm$0.52}  & -4.38{\scriptsize$\pm$0.45}  \\
\hline
\multirow{3}{*}{\shortstack{Llama-3.1\\-8B}} & Vul. & 3.89{\scriptsize$\pm$1.39} & 4.63{\scriptsize$\pm$1.39} & 1.97{\scriptsize$\pm$1.44} \\ 
& Susc. & 1.30{\scriptsize$\pm$1.43} & 2.00{\scriptsize$\pm$1.41} & 1.38{\scriptsize$\pm$1.49} \\
& Random & -0.93{\scriptsize$\pm$0.41} & -2.56{\scriptsize$\pm$0.36} & -4.92{\scriptsize$\pm$0.53} \\
\hline
\end{tabular}
\vspace{-0.4cm}
\end{table}

\section{Discussion}

\subsection{Generalizability of Identified Heads}

\begin{figure}
\vspace{-0.0cm}
\centering
\includegraphics[width=0.95\linewidth]{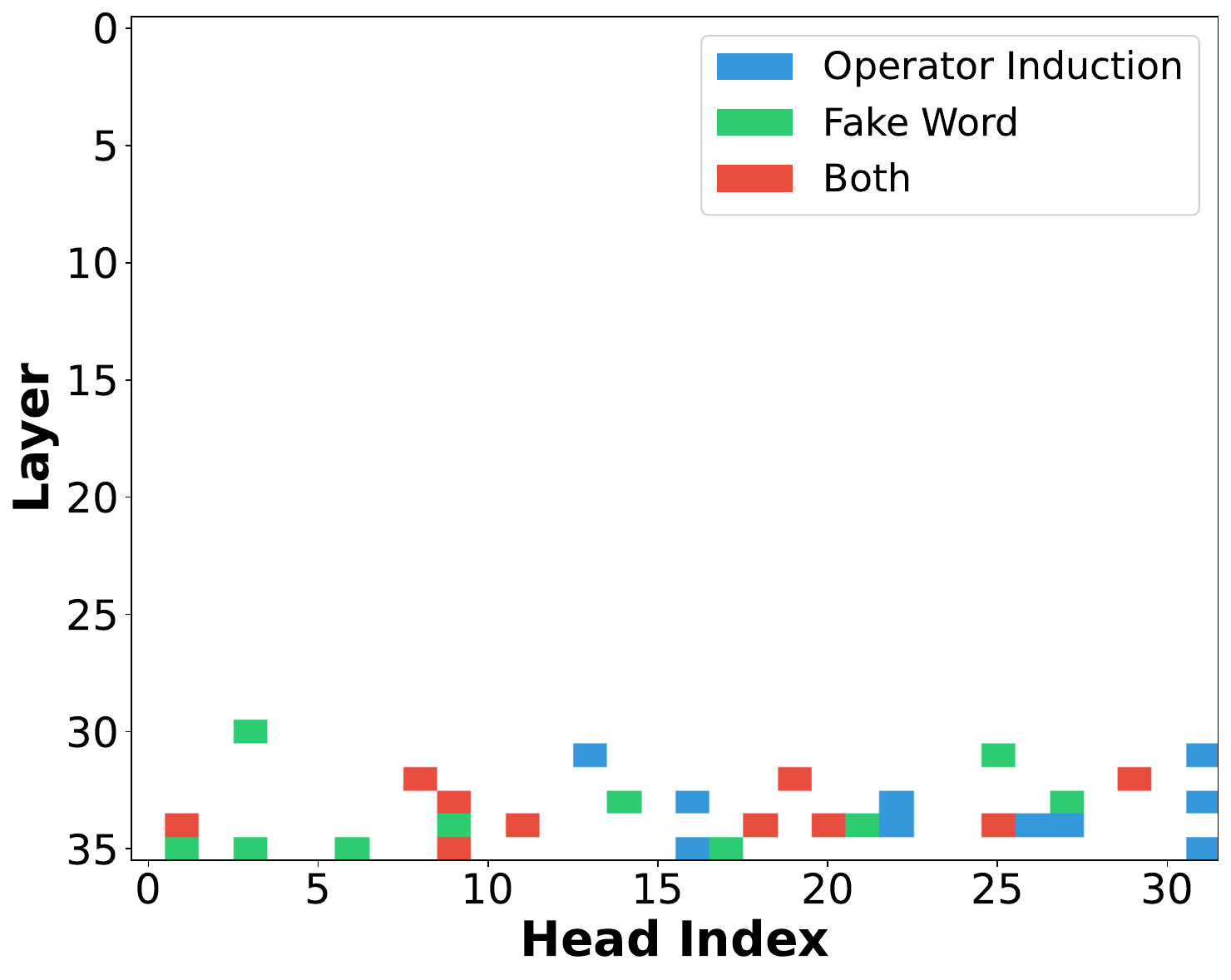}
\caption{Cross-task overlap of Susceptibility Heads between Operator Induction and Fake Word Inference tasks.}
\label{fig:overlapping}
\vspace{-0.55cm}
\end{figure}

To evaluate whether identified heads capture generalizable conflict-resolution mechanisms, we measure the consistency of both Vulnerability and Susceptible Heads across the Operator Induction and Fake Word Inference task. On the first hand, we find that susceptible heads exhibit significant overlap across tasks, with 10 of the top 20 heads from Qwen3-4B shared between both tasks, as visualized in Figure~\ref{fig:overlapping}. This observation suggests that late-layer conflict resolution relies on generalizable mechanisms. To verify the generalizability, we conduct a cross-task ablation study: ablating top 10 susceptibility heads identified from Fake Word Inference for Operator Induction. This cross-task ablation improves performance under corruption by 3.80\% in average across models, indicating a potential mitigation solution under conflict demonstrations across different rule inference tasks. On the other hand, we observe that vulnerability heads exhibit lower cross-task overlap (4 of top 20 heads from Qwen3-4B), suggesting that attention allocation and sensitivity to conflicting evidence is more task-specific.

\subsection{Empirical Synergy between the Two Phases}

We also investigate in the synergy among these components within the model's computational pathway. Specifically, we examine whether ablating vulnerability heads reduces the susceptibility of late-layer heads by measuring changes in their logit contributions, and compare these effects against random head ablations. As shown in Figure \ref{fig:synergy} at Appendix \ref{app:additional}, ablating vulnerability heads reduces the susceptibility scores of top susceptible heads (e.g. L30H29 and L31H5 for Qwen3-4B on Operator Induction) significantly more than random ablation. This case study suggests an empirical synergy between intermediate-layer vulnerability creation and late-layer susceptibility, through the mediation of the two types of heads.

\subsection{Vulnerability Heads are Relevant to Positional Bias}

Beyond performance improvement, we also notice that ablating vulnerability heads reduces positional bias in corruption sensitivity. We formally define positional bias as $\text{PB} = \text{Var}_{p=1}^k[\Delta\text{Acc}_p]$, where $\Delta\text{Acc}_p$ represents the accuracy decrease when position $p$ is corrupted. As shown in Table \ref{tab:variance_reduction} at Appendix \ref{app:additional}, removing vulnerability heads reduces positional variance moderately across models. Since vulnerability heads exhibit disproportionate attention to certain positions while showing high sensitivity to corruption at those positions, their removal likely mitigates the uneven impact of corruption across demonstration sequences, leading to more uniform vulnerability patterns.


\section{Conclusion}

In this work, we provide a mechanistic analysis of how LLMs process conflicting demonstrations during in-context rule inference. Employing two tasks with strict demonstration reliance, we uncover a two-phase computational structure where models encode both correct and corrupted rules in intermediate layers before resolving them in late layers. We then identify and localize two distinct types of attention heads: vulnerability heads that exhibit positional bias and high sensitivity to corruption, and susceptible heads that reduce support for correct predictions when exposed to conflicting evidence. Our ablation results show that masking them mitigates performance degradation by over 10\%, confirming the causal role of these identified attention heads.

\clearpage
\section*{Impact Statement}

This paper presents mechanistic interpretability research aimed at understanding how large language models process conflicting demonstrations during in-context learning. Our work identifies specific failure modes and model components responsible for reasoning errors under contradictory information, which could benefit multiple stakeholders. For researchers and practitioners, our findings could inform the design of more robust prompting strategies, help detect when models are being misled by noisy or adversarial demonstrations, and guide the development of architectures more resilient to conflicting evidence. The identified attention heads and ablation methods provide insights for improving model reliability in practical applications where demonstration quality cannot be guaranteed.

While our research primarily advances understanding of model internals for safety and robustness purposes, we acknowledge that detailed knowledge of vulnerability mechanisms could potentially inform adversarial attack strategies. However, we believe the benefits of transparency about model failure modes substantially outweigh such risks, as understanding vulnerabilities is a prerequisite for developing effective defenses. Our work contributes to the broader goal of making language models more reliable and trustworthy, particularly in high-stakes applications where reasoning under uncertainty and conflicting information is critical.




\bibliography{example_paper}

@article{brown2020language,
  title={Language models are few-shot learners},
  author={Brown, Tom and Mann, Benjamin and Ryder, Nick and Subbiah, Melanie and Kaplan, Jared D and Dhariwal, Prafulla and Neelakantan, Arvind and Shyam, Pranav and Sastry, Girish and Askell, Amanda and others},
  journal={Advances in neural information processing systems},
  volume={33},
  pages={1877--1901},
  year={2020}
}

@article{dong2022survey,
  title={A survey on in-context learning},
  author={Dong, Qingxiu and Li, Lei and Dai, Damai and Zheng, Ce and Ma, Jingyuan and Li, Rui and Xia, Heming and Xu, Jingjing and Wu, Zhiyong and Liu, Tianyu and others},
  journal={arXiv preprint arXiv:2301.00234},
  year={2022}
}

@article{xu2024knowledge,
  title={Knowledge conflicts for llms: A survey},
  author={Xu, Rongwu and Qi, Zehan and Guo, Zhijiang and Wang, Cunxiang and Wang, Hongru and Zhang, Yue and Xu, Wei},
  journal={arXiv preprint arXiv:2403.08319},
  year={2024}
}

@article{zhang2025understanding,
  title={Understanding and mitigating political stance cross-topic generalization in large language models},
  author={Zhang, Jiayi and Yang, Shu and Wu, Junchao and Wong, Derek F and Wang, Di},
  journal={arXiv preprint arXiv:2508.02360},
  year={2025}
}

@article{fu2025short,
  title={Short-length Adversarial Training Helps LLMs Defend Long-length Jailbreak Attacks: Theoretical and Empirical Evidence},
  author={Fu, Shaopeng and Ding, Liang and Zhang, Jingfeng and Wang, Di},
  journal={arXiv preprint arXiv:2502.04204},
  year={2025}
}

@article{su2025understanding,
  title={Understanding how value neurons shape the generation of specified values in llms},
  author={Su, Yi and Zhang, Jiayi and Yang, Shu and Wang, Xinhai and Hu, Lijie and Wang, Di},
  journal={arXiv preprint arXiv:2505.17712},
  year={2025}
}

@article{dong2025understanding,
  title={Understanding and Mitigating Cross-lingual Privacy Leakage via Language-specific and Universal Privacy Neurons},
  author={Dong, Wenshuo and Yang, Qingsong and Yang, Shu and Hu, Lijie and Ding, Meng and Lin, Wanyu and Zheng, Tianhang and Wang, Di},
  journal={arXiv preprint arXiv:2506.00759},
  year={2025}
}

@article{yang2024exploring,
  title={Exploring the Personality Traits of LLMs through Latent Features Steering},
  author={Yang, Shu and Zhu, Shenzhe and Liu, Liang and Hu, Lijie and Li, Mengdi and Wang, Di},
  journal={arXiv preprint arXiv:2410.10863},
  year={2024}
}

@article{hu2024hopfieldian,
  title={A hopfieldian view-based interpretation for chain-of-thought reasoning},
  author={Hu, Lijie and Liu, Liang and Yang, Shu and Chen, Xin and Xiao, Hongru and Li, Mengdi and Zhou, Pan and Ali, Muhammad Asif and Wang, Di},
  journal={arXiv preprint arXiv:2406.12255},
  year={2024}
}

@article{cheng2024multi,
  title={Multi-hop question answering under temporal knowledge editing},
  author={Cheng, Keyuan and Lin, Gang and Fei, Haoyang and Yu, Lu and Ali, Muhammad Asif and Hu, Lijie and Wang, Di and others},
  journal={arXiv preprint arXiv:2404.00492},
  year={2024}
}

@article{zhang2024locate,
  title={Locate-then-edit for multi-hop factual recall under knowledge editing},
  author={Zhang, Zhuoran and Li, Yongxiang and Kan, Zijian and Cheng, Keyuan and Hu, Lijie and Wang, Di},
  journal={arXiv preprint arXiv:2410.06331},
  year={2024}
}

@article{zhang2025modalities,
  title={When modalities conflict: How unimodal reasoning uncertainty governs preference dynamics in mllms},
  author={Zhang, Zhuoran and Wang, Tengyue and Gong, Xilin and Shi, Yang and Wang, Haotian and Wang, Di and Hu, Lijie},
  journal={arXiv preprint arXiv:2511.02243},
  year={2025}
}

@inproceedings{xie2023adaptive,
  title={Adaptive chameleon or stubborn sloth: Revealing the behavior of large language models in knowledge conflicts},
  author={Xie, Jian and Zhang, Kai and Chen, Jiangjie and Lou, Renze and Su, Yu},
  booktitle={The Twelfth International Conference on Learning Representations},
  year={2023}
}

@article{zhang2025eap,
  title={Eap-gp: Mitigating saturation effect in gradient-based automated circuit identification},
  author={Zhang, Lin and Dong, Wenshuo and Zhang, Zhuoran and Yang, Shu and Hu, Lijie and Liu, Ninghao and Zhou, Pan and Wang, Di},
  journal={arXiv preprint arXiv:2502.06852},
  year={2025}
}

@article{li2023contradoc,
  title={Contradoc: understanding self-contradictions in documents with large language models},
  author={Li, Jierui and Raheja, Vipul and Kumar, Dhruv},
  journal={arXiv preprint arXiv:2311.09182},
  year={2023}
}

@article{wang2024eliminating,
  title={Eliminating position bias of language models: A mechanistic approach},
  author={Wang, Ziqi and Zhang, Hanlin and Li, Xiner and Huang, Kuan-Hao and Han, Chi and Ji, Shuiwang and Kakade, Sham M and Peng, Hao and Ji, Heng},
  journal={arXiv preprint arXiv:2407.01100},
  year={2024}
}

@article{min2022rethinking,
  title={Rethinking the role of demonstrations: What makes in-context learning work?},
  author={Min, Sewon and Lyu, Xinxi and Holtzman, Ari and Artetxe, Mikel and Lewis, Mike and Hajishirzi, Hannaneh and Zettlemoyer, Luke},
  journal={arXiv preprint arXiv:2202.12837},
  year={2022}
}

@article{yoo2022ground,
  title={Ground-truth labels matter: A deeper look into input-label demonstrations},
  author={Yoo, Kang Min and Kim, Junyeob and Kim, Hyuhng Joon and Cho, Hyunsoo and Jo, Hwiyeol and Lee, Sang-Woo and Lee, Sang-goo and Kim, Taeuk},
  journal={arXiv preprint arXiv:2205.12685},
  year={2022}
}

@article{wei2023larger,
  title={Larger language models do in-context learning differently},
  author={Wei, Jerry and Wei, Jason and Tay, Yi and Tran, Dustin and Webson, Albert and Lu, Yifeng and Chen, Xinyun and Liu, Hanxiao and Huang, Da and Zhou, Denny and others},
  journal={arXiv preprint arXiv:2303.03846},
  year={2023}
}

@article{halawi2023overthinking,
  title={Overthinking the truth: Understanding how language models process false demonstrations},
  author={Halawi, Danny and Denain, Jean-Stanislas and Steinhardt, Jacob},
  journal={arXiv preprint arXiv:2307.09476},
  year={2023}
}

@article{yu2024large,
  title={How do large language models learn in-context? query and key matrices of in-context heads are two towers for metric learning},
  author={Yu, Zeping and Ananiadou, Sophia},
  journal={arXiv preprint arXiv:2402.02872},
  year={2024}
}

@article{hua2025vision,
  title={How Do Vision-Language Models Process Conflicting Information Across Modalities?},
  author={Hua, Tianze and Yun, Tian and Pavlick, Ellie},
  journal={arXiv preprint arXiv:2507.01790},
  year={2025}
}

@article{jiao2023spin,
  title={SPIN: Sparsifying and Integrating Internal Neurons in Large Language Models for Text Classification},
  author={Jiao, Difan and Liu, Yilun and Tang, Zhenwei and Matter, Daniel and Pfeffer, J{\"u}rgen and Anderson, Ashton},
  journal={arXiv preprint arXiv:2311.15983},
  year={2023}
}

@inproceedings{yang2022empirical,
  title={An empirical study of gpt-3 for few-shot knowledge-based vqa},
  author={Yang, Zhengyuan and Gan, Zhe and Wang, Jianfeng and Hu, Xiaowei and Lu, Yumao and Liu, Zicheng and Wang, Lijuan},
  booktitle={Proceedings of the AAAI conference on artificial intelligence},
  volume={36},
  number={3},
  pages={3081--3089},
  year={2022}
}

@article{wei2021finetuned,
  title={Finetuned language models are zero-shot learners},
  author={Wei, Jason and Bosma, Maarten and Zhao, Vincent Y and Guu, Kelvin and Yu, Adams Wei and Lester, Brian and Du, Nan and Dai, Andrew M and Le, Quoc V},
  journal={arXiv preprint arXiv:2109.01652},
  year={2021}
}

@article{zong2024vl,
  title={Vl-icl bench: The devil in the details of benchmarking multimodal in-context learning},
  author={Zong, Yongshuo and Bohdal, Ondrej and Hospedales, Timothy M},
  journal={CoRR},
  year={2024}
}

@article{yang2025qwen3,
  title={Qwen3 technical report},
  author={Yang, An and Li, Anfeng and Yang, Baosong and Zhang, Beichen and Hui, Binyuan and Zheng, Bo and Yu, Bowen and Gao, Chang and Huang, Chengen and Lv, Chenxu and others},
  journal={arXiv preprint arXiv:2505.09388},
  year={2025}
}

@article{dubey2024llama,
  title={The llama 3 herd of models},
  author={Dubey, Abhimanyu and Jauhri, Abhinav and Pandey, Abhinav and Kadian, Abhishek and Al-Dahle, Ahmad and Letman, Aiesha and Mathur, Akhil and Schelten, Alan and Yang, Amy and Fan, Angela and others},
  journal={arXiv e-prints},
  pages={arXiv--2407},
  year={2024}
}

@article{alain2016understanding,
  title={Understanding intermediate layers using linear classifier probes},
  author={Alain, Guillaume and Bengio, Yoshua},
  journal={arXiv preprint arXiv:1610.01644},
  year={2016}
}

@article{elazar2021amnesic,
  title={Amnesic probing: Behavioral explanation with amnesic counterfactuals},
  author={Elazar, Yanai and Ravfogel, Shauli and Jacovi, Alon and Goldberg, Yoav},
  journal={Transactions of the Association for Computational Linguistics},
  volume={9},
  pages={160--175},
  year={2021},
  publisher={MIT Press One Rogers Street, Cambridge, MA 02142-1209, USA journals-info~…}
}

@misc{nostalgebraist2020logitlens,
  author       = {nostalgebraist},
  title        = {Interpreting GPT: The Logit Lens},
  year         = {2020},
  howpublished = {LessWrong post},
  note         = {\url{https://www.lesswrong.com/posts/AcKRB8wDpdaN6v6ru/interpreting-gpt-the-logit-lens}},
}

@article{cobbina2025show,
  title={Where to show demos in your prompt: A positional bias of in-context learning},
  author={Cobbina, Kwesi and Zhou, Tianyi},
  journal={arXiv preprint arXiv:2507.22887},
  year={2025}
}

@article{hsieh2024found,
  title={Found in the middle: Calibrating positional attention bias improves long context utilization},
  author={Hsieh, Cheng-Yu and Chuang, Yung-Sung and Li, Chun-Liang and Wang, Zifeng and Le, Long T and Kumar, Abhishek and Glass, James and Ratner, Alexander and Lee, Chen-Yu and Krishna, Ranjay and others},
  journal={arXiv preprint arXiv:2406.16008},
  year={2024}
}

@article{peysakhovich2023attention,
  title={Attention Sorting Combats Recency Bias In Long Context Language Models},
  author={Peysakhovich, Alexander and Bastani, Osbert},
  journal={arXiv preprint arXiv:2310.01427},
  year={2023}
}

@article{chen2023positional,
  title={Positional Information Matters for Invariant In-Context Learning: A Case Study of Simple Function Classes},
  author={Chen, Yongqiang and others},
  journal={arXiv preprint arXiv:2311.18194},
  year={2023}
}

@article{han2024unibias,
  title={UniBias: Unveiling and Mitigating LLM Bias through Internal Attention and FFN Manipulation},
  author={Han, Hanzhang and others},
  journal={arXiv preprint arXiv:2405.20612},
  year={2024}
}

@article{cho2024revisiting,
  title={Revisiting in-context learning inference circuit in large language models},
  author={Cho, Hakaze and Kato, Mariko and Sakai, Yoshihiro and Inoue, Naoya},
  journal={arXiv preprint arXiv:2410.04468},
  year={2024}
}

@article{wang2022interpretability,
  title={Interpretability in the wild: a circuit for indirect object identification in gpt-2 small},
  author={Wang, Kevin and Variengien, Alexandre and Conmy, Arthur and Shlegeris, Buck and Steinhardt, Jacob},
  journal={arXiv preprint arXiv:2211.00593},
  year={2022}
}

@article{olsson2022context,
  title={In-context learning and induction heads},
  author={Olsson, Catherine and Elhage, Nelson and Nanda, Neel and Joseph, Nicholas and DasSarma, Nova and Henighan, Tom and Mann, Ben and Askell, Amanda and Bai, Yuntao and Chen, Anna and others},
  journal={arXiv preprint arXiv:2209.11895},
  year={2022}
}

@article{meng2022locating,
  title={Locating and editing factual associations in gpt},
  author={Meng, Kevin and Bau, David and Andonian, Alex and Belinkov, Yonatan},
  journal={Advances in neural information processing systems},
  volume={35},
  pages={17359--17372},
  year={2022}
}

@article{zhang2023towards,
  title={Towards best practices of activation patching in language models: Metrics and methods},
  author={Zhang, Fred and Nanda, Neel},
  journal={arXiv preprint arXiv:2309.16042},
  year={2023}
}

@article{kossen2023context,
  title={In-context learning learns label relationships but is not conventional learning},
  author={Kossen, Jannik and Gal, Yarin and Rainforth, Tom},
  journal={arXiv preprint arXiv:2307.12375},
  year={2023}
}

@article{zhang2024trained,
  title={Trained transformers learn linear models in-context},
  author={Zhang, Ruiqi and Frei, Spencer and Bartlett, Peter L},
  journal={Journal of Machine Learning Research},
  volume={25},
  number={49},
  pages={1--55},
  year={2024}
}

@article{coda2023meta,
  title={Meta-in-context learning in large language models},
  author={Coda-Forno, Julian and Binz, Marcel and Akata, Zeynep and Botvinick, Matt and Wang, Jane and Schulz, Eric},
  journal={Advances in Neural Information Processing Systems},
  volume={36},
  pages={65189--65201},
  year={2023}
}

@article{pan2021attacking,
  title={Attacking open-domain question answering by injecting misinformation},
  author={Pan, Liangming and Chen, Wenhu and Kan, Min-Yen and Wang, William Yang},
  journal={arXiv preprint arXiv:2110.07803},
  year={2021}
}

@inproceedings{von2023transformers,
  title={Transformers learn in-context by gradient descent},
  author={Von Oswald, Johannes and Niklasson, Eyvind and Randazzo, Ettore and Sacramento, Jo{\~a}o and Mordvintsev, Alexander and Zhmoginov, Andrey and Vladymyrov, Max},
  booktitle={International Conference on Machine Learning},
  pages={35151--35174},
  year={2023},
  organization={PMLR}
}

@article{bai2023transformers,
  title={Transformers as statisticians: Provable in-context learning with in-context algorithm selection},
  author={Bai, Yu and Chen, Fan and Wang, Huan and Xiong, Caiming and Mei, Song},
  journal={Advances in neural information processing systems},
  volume={36},
  pages={57125--57211},
  year={2023}
}

@article{cammarata2021curve,
  title={Curve circuits},
  author={Cammarata, Nick and Goh, Gabriel and Carter, Shan and Voss, Chelsea and Schubert, Ludwig and Olah, Chris},
  journal={Distill},
  volume={6},
  number={1},
  pages={e00024--006},
  year={2021}
}

@misc{elhage2021framework,
  title        = {A Mathematical Framework for Transformer Circuits},
  author       = {Elhage, Nelson and Nanda, Neel and Olsson, Catherine and Henighan, Tom and Joseph, Nicholas and Mann, Ben and Askell, Amanda and Bai, Yuntao and Chen, Anna and Conerly, Tom and DasSarma, Nova and Drain, Dawn and Ganguli, Deep and Hatfield-Dodds, Zac and Hernandez, Danny and Jones, Andy and Kernion, Jackson and Lovitt, Liane and Ndousse, Kamal and Amodei, Dario and Brown, Tom and Clark, Jack and Kaplan, Jared and McCandlish, Sam and Olah, Chris},
  organization = {Anthropic},
  year         = {2021},
  month        = dec,
  day          = {22},
  url          = {https://transformer-circuits.pub/2021/framework/index.html},
  note         = {Transformer Circuits series}
}

@article{olah2020zoom,
  title   = {Zoom In: An Introduction to Circuits},
  author  = {Olah, Chris and Cammarata, Nick and Schubert, Ludwig and Goh, Gabriel and Petrov, Michael and Carter, Shan},
  journal = {Distill},
  year    = {2020},
  month   = mar,
  day     = {10},
  doi     = {10.23915/distill.00024.001},
  url     = {https://distill.pub/2020/circuits/zoom-in},
  note    = {OpenAI}
}

@inproceedings{kwon2023efficient,
  title={Efficient memory management for large language model serving with pagedattention},
  author={Kwon, Woosuk and Li, Zhuohan and Zhuang, Siyuan and Sheng, Ying and Zheng, Lianmin and Yu, Cody Hao and Gonzalez, Joseph and Zhang, Hao and Stoica, Ion},
  booktitle={Proceedings of the 29th symposium on operating systems principles},
  pages={611--626},
  year={2023}
}

@inproceedings{wendler2024llamas,
  title={Do llamas work in english? on the latent language of multilingual transformers},
  author={Wendler, Chris and Veselovsky, Veniamin and Monea, Giovanni and West, Robert},
  booktitle={Proceedings of the 62nd Annual Meeting of the Association for Computational Linguistics (Volume 1: Long Papers)},
  pages={15366--15394},
  year={2024}
}

@article{fu2025deep,
  title={Deep think with confidence},
  author={Fu, Yichao and Wang, Xuewei and Tian, Yuandong and Zhao, Jiawei},
  journal={arXiv preprint arXiv:2508.15260},
  year={2025}
}

@article{kang2025scalable,
  title={Scalable best-of-n selection for large language models via self-certainty},
  author={Kang, Zhewei and Zhao, Xuandong and Song, Dawn},
  journal={arXiv preprint arXiv:2502.18581},
  year={2025}
}

@article{zur2025language,
  title={Are language models aware of the road not taken? Token-level uncertainty and hidden state dynamics},
  author={Zur, Amir and Geiger, Atticus and Lubana, Ekdeep Singh and Bigelow, Eric},
  journal={arXiv preprint arXiv:2511.04527},
  year={2025}
}

@article{wang2023adversarial,
  title={Adversarial demonstration attacks on large language models},
  author={Wang, Jiongxiao and Liu, Zichen and Park, Keun Hee and Jiang, Zhuojun and Zheng, Zhaoheng and Wu, Zhuofeng and Chen, Muhao and Xiao, Chaowei},
  journal={arXiv preprint arXiv:2305.14950},
  year={2023}
}

@inproceedings{li2024debiasing,
  title={Debiasing in-context learning by instructing LLMs how to follow demonstrations},
  author={Li, Lvxue and Chen, Jiaqi and Lu, Xinyu and Lu, Yaojie and Lin, Hongyu and Zhou, Shuheng and Zhu, Huijia and Wang, Weiqiang and Liu, Zhongyi and Han, Xianpei and others},
  booktitle={Findings of the Association for Computational Linguistics ACL 2024},
  pages={7203--7215},
  year={2024}
}

@article{wang2023label,
  title={Label words are anchors: An information flow perspective for understanding in-context learning},
  author={Wang, Lean and Li, Lei and Dai, Damai and Chen, Deli and Zhou, Hao and Meng, Fandong and Zhou, Jie and Sun, Xu},
  journal={arXiv preprint arXiv:2305.14160},
  year={2023}
}

@inproceedings{qin2024context,
  title={In-context learning with iterative demonstration selection},
  author={Qin, Chengwei and Zhang, Aston and Chen, Chen and Dagar, Anirudh and Ye, Wenming},
  booktitle={Findings of the Association for Computational Linguistics: EMNLP 2024},
  pages={7441--7455},
  year={2024}
}

@article{hu2024strategic,
  title={Strategic demonstration selection for improved fairness in LLM in-context learning},
  author={Hu, Jingyu and Liu, Weiru and Du, Mengnan},
  journal={arXiv preprint arXiv:2408.09757},
  year={2024}
}

@article{liu2023towards,
  title={Towards understanding in-context learning with contrastive demonstrations and saliency maps},
  author={Liu, Fuxiao and Xu, Paiheng and Li, Zongxia and Feng, Yue and Song, Hyemi},
  journal={arXiv preprint arXiv:2307.05052},
  year={2023}
}
\bibliographystyle{icml2026}

\newpage
\appendix
\onecolumn

\section{Details of Corruption-based Intervention Framework}
\label{app:framework}

\subsection{Corruption-based Experimental Design}

Our evaluation protocol systematically examines model robustness to conflicting demonstrations through controlled corruption at specific positions within the demonstration sequence.

\subsubsection{Data Scale and Coverage}
Taking 4-shot ICL as an example, we evaluate the following:

\xhdr{Operator Induction Tasks}
\begin{itemize}
    \item Query samples: 60 unique queries per task
    \item Shots per query: 4 demonstrations
    \item Corruption positions: 4 positions
    \item Rollouts per position: 3 independent trials with different random seeds
    \item Total evaluations: 720 comparisons (60 × 4 × 3)
\end{itemize}

\xhdr{Fake Word Inference Task}
\begin{itemize}
    \item Query samples: 100 unique target pairs (color-object combinations)
    \item Shots per query: 4 demonstrations
    \item Corruption positions: 4 positions
    \item Rollouts per position: 3 independent trials with different random seeds
    \item Total evaluations: 1,200 comparisons (100 × 4 × 3)
\end{itemize}

Note that in our tasks the choice of corrupted rules can be flexible based on the hypothesis rule space. This design yields robust statistical estimates across position-specific corruption patterns. Table \ref{tab:baseline_performance} validates our task selection criteria for studying demonstration-dependent conflict resolution. All tasks exhibit near-chance 0-shot performance, confirming genuine demonstration reliance. Performance substantially improves with few-shot demonstration, demonstrating reliable ICL capability. This pattern of models' failing without demonstrations yet succeeding with them intrinsically satisfies our principled framework requirement that models must genuinely rely on contextual evidence rather than parametric knowledge.

\begin{table}[h]
\centering
\caption{Clean (Uncorrupted) in-context learning performance across tasks and models. Models exhibit reliable ICL capabilities on both Operator Induction and Fake Word Inference.}
\label{tab:baseline_performance}
\begin{tabular}{llcccc}
\toprule
\textbf{Task} & \textbf{Model} & \textbf{0-shot} & \textbf{4-shot} & \textbf{6-shot} & \textbf{8-shot} \\
\midrule
\multirow{4}{*}{\textbf{Operator Induction}} 
& Qwen3-0.6B & 0.35 & 0.84{\scriptsize$\pm$0.03} & 0.88{\scriptsize$\pm$0.01} & 0.88{\scriptsize$\pm$0.01} \\
& Qwen3-4B & 0.37 & 0.99{\scriptsize$\pm$0.01} & 1.00{\scriptsize$\pm$0.00} & 1.00{\scriptsize$\pm$0.00} \\
& Llama-3.2-3B & 0.42 & 0.83{\scriptsize$\pm$0.05} & 0.69{\scriptsize$\pm$0.08} & 0.75{\scriptsize$\pm$0.02} \\
& Llama-3.1-8B & 0.32 & 0.90{\scriptsize$\pm$0.02} & 0.92{\scriptsize$\pm$0.02} & 0.91{\scriptsize$\pm$0.01} \\
\midrule
\multirow{4}{*}{\textbf{Fake Word Inference}} 
& Qwen3-0.6B & 0.03 & 0.95{\scriptsize$\pm$0.02} & 0.88{\scriptsize$\pm$0.01} & 0.91{\scriptsize$\pm$0.02} \\
& Qwen3-4B & 0.03 & 0.99{\scriptsize$\pm$0.01} & 1.00{\scriptsize$\pm$0.00} & 1.00{\scriptsize$\pm$0.00} \\
& Llama-3.2-3B & 0.07 & 0.82{\scriptsize$\pm$0.01} & 0.84{\scriptsize$\pm$0.02} & 0.84{\scriptsize$\pm$0.02} \\
& Llama-3.1-8B & 0.08 & 0.93{\scriptsize$\pm$0.01} & 0.93{\scriptsize$\pm$0.01} & 0.93{\scriptsize$\pm$0.01} \\
\bottomrule
\end{tabular}
\end{table}

\subsubsection{Corruption Mechanism}
We implement systematic single-position corruption where exactly one demonstration at position $i$ (e.g., $i \in \{0, 1, 2, 3\}$ for 4-shot scenarios) receives a corrupted rule while maintaining the majority rule principle. For Operator Induction tasks, corruption involves replacing the correct mathematical operator with an alternative operator (+, -, ×), ensuring the corrupted demonstration exhibits a different underlying rule while maintaining surface-level plausibility. For Fake Word Inference, we first select all demonstrations showing the correct fake word mapping (e.g., all 4 demonstrations mapping the fake color word to the target real color), then substitute exactly one demonstration's real word with an incorrect alternative from the same category (e.g., replacing ``purple'' with ``blue'' while keeping the fake word unchanged). This ensures the correct mapping maintains a 3:1 majority over the corrupted alternative, providing dominating evidence for conflict resolution while testing whether models will be obfuscated by the only corrupted evidence.

\subsection{Model Response Evaluation Protocol}

\subsubsection{Inference Setup}

We employ the vLLM framework \citep{kwon2023efficient} for efficient batched inference on NVIDIA A100 80GB GPUs. For generation, we use greedy decoding (temperature = 0.0) with a maximum output length of 2,048 tokens. For Qwen3 models, we enable the extended thinking mode via the chat template to allow chain-of-thought reasoning.

\subsubsection{Prompt Design}

\xhdr{Operator Induction}
Taking 4-shot in-context rule inference as an example, we employ standard formatting adapted from \citep{zong2024vl}. The prompt instructs models to output their final answer after the \texttt{\#\#\#\#} delimiter. Below is an example of corrupted prompt:

\begin{verbatim}
The text contains two digit numbers and a ? representing the 
mathematical operator. Induce the mathematical operator (addition, 
multiplication, minus) according to the results of the in-context 
examples and calculate the result. Think step by step, then write 
your final answer after ####

Support Set:
8 ? 6 =
Answer: 14

3 ? 5 =
Answer: 8

7 ? 2 = 
Answer: 5

9 ? 4 = 
Answer: 13

Question:
6 ? 3 = ?

Answer:
\end{verbatim}

\xhdr{Fake Word Inference}
For the Fake Word Inference task, we present demonstrations mapping fake words to real color-object combinations. Each demonstration follows the format ``\texttt{[fake\_phrase] means [real\_phrase]}''. In this corrupted example, one demonstration (position 1) shows an incorrect color mapping while the majority maintain the correct mapping. Below is an example of corrupted prompt:

\begin{verbatim}
You will see examples showing what fake words mean in terms of 
real colors and objects. Learn the mapping from fake words to 
real words, then answer the question using real words only. 
Think step by step, then write your final answer after ####

Support Set:
zandolex vundelka means purple sheep
zandolex glemorax means blue hat
zandolex plintovar means purple scarf
zandolex thovaline means purple flower

Question:
What color is zandolex frivenmox?

Answer:
\end{verbatim}

\subsubsection{Answer Extraction and Evaluation}
\label{app:main_eval}

\xhdr{Operator Induction Evaluation}
For Operator Induction tasks, we employ a priority-based regex extraction approach to identify the final numerical answer from model responses. The extraction follows this priority order: (1) numbers appearing after the \texttt{\#\#\#\#} delimiter; (2) numbers after explicit answer markers such as ``Final answer:'' or ``Answer:''; (3) numbers immediately following an equals sign; and (4) the last number in the response as a fallback. Correctness is determined by exact match between the extracted integer and the ground truth answer.

\xhdr{Fake Word Inference Evaluation}
For Fake Word Inference, we extract the final answer segment after the \texttt{\#\#\#\#} delimiter and perform exact string matching against the ground truth color (e.g., ``purple''). This deterministic evaluation ensures consistent assessment of mapping accuracy.

\subsubsection{Statistical Design}

\xhdr{Majority Rule Verification}
Our corruption mechanism maintains the majority rule principle: in 4-shot scenarios with single-position corruption, correct demonstrations maintain at least a 3:1 majority over corrupted ones, providing a theoretically sound basis for expecting successful conflict resolution under ideal conditions. This applies to both Operator Induction tasks (correct operator appears in 3 of 4 demonstrations) and Fake Word Inference (correct mapping appears in 3 of 4 examples).

\xhdr{Paired Experimental Control} 
Critical to our methodology is the paired nature of evaluations: for each (query, position, rollout) tuple, the clean baseline and corrupted conditions share the same underlying demonstration structure, differing only in the corruption applied at the target position. We use position-specific random seeds to ensure reproducibility across experimental runs. We report performance degradation as the difference between clean baseline accuracy and corrupted accuracy, with statistical significance assessed via paired $t$-tests across rollouts.

\begin{figure}[t!]
    \centering
    \includegraphics[width=0.99\linewidth]{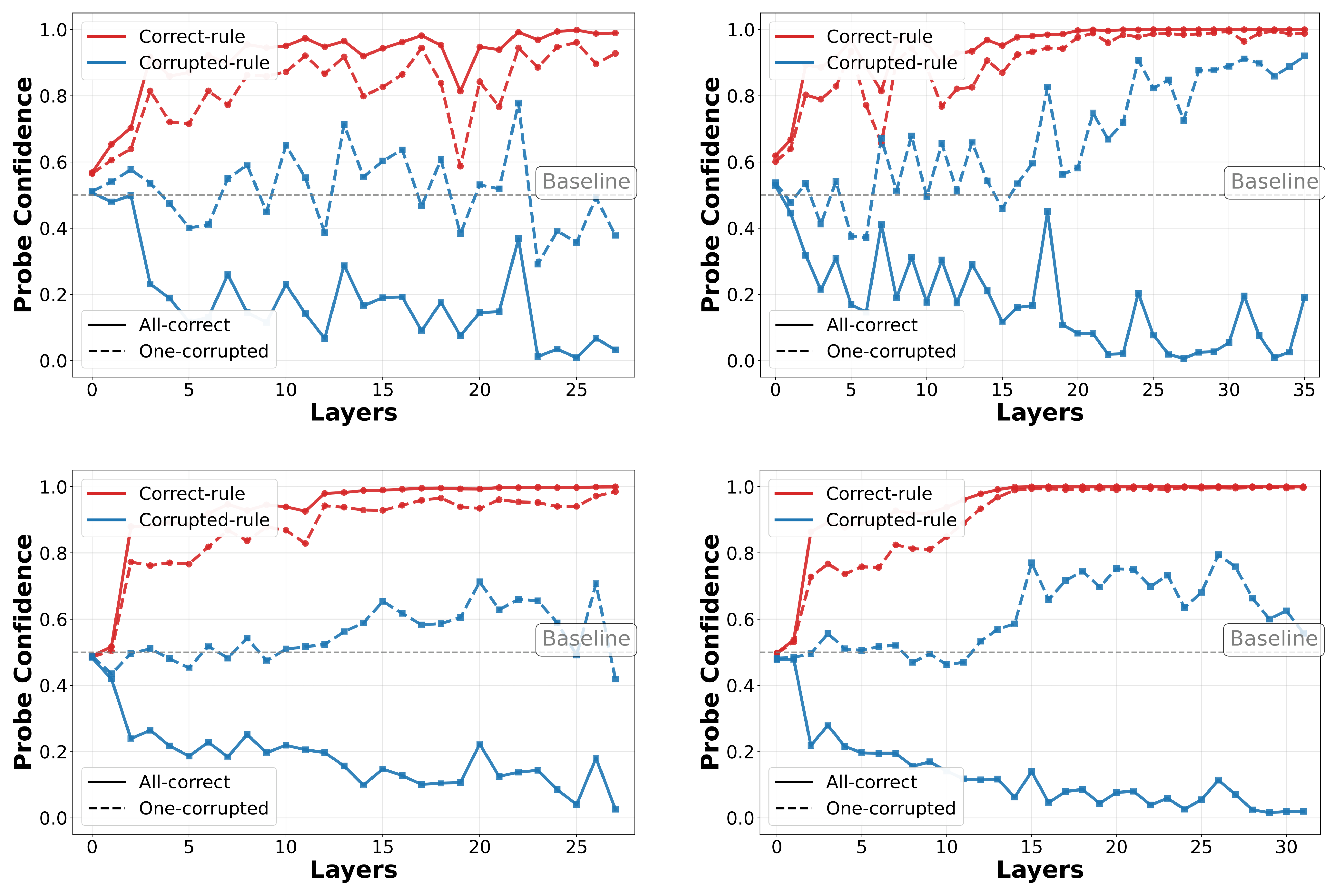}
    \caption{Linear probe confidence across model layers for Qwen3-0.6B, Qwen3-4B, Llama-3.1-8B-Instruct, and Llama-3.2-3B-Instruct from left to right, up to down. \textbf{Red solid}: Correct-rule probe confidence in all-correct scenarios. \textbf{Red dashed}: Correct-rule probe confidence in one-corrupted scenarios. \textbf{Blue solid}: Corrupted-rule probe confidence in all-correct scenarios. \textbf{Blue dashed}: Corrupted-rule probe confidence in one-corrupted scenarios. The baseline represents chance-level performance.}
    \label{fig:probe_analysis_detailed}
    \vspace{-0.6cm}
\end{figure}

\section{Linear Probe Analysis Details}
\label{app:linear_probes}

\subsection{Probe Training Methodology}
We train linear probes as binary classifiers to detect the presence of specific mathematical operators (+, -, ×) in model internal representations. Each probe learns to answer ``Is the target rule present in the demonstrations?'' rather than performing multi-class operator identification. This design provides a principled approach to measure rule encoding strength across different corruption scenarios.

For each target operator, we generate 600 balanced training samples, containing 300 positive samples for the target rule. Take the Operator Induction task under 4-shot ICL as an example, we compose the positive samples 75 each with 1/4, 2/4, 3/4, 4/4 target demonstrations, and target rules are completely absent in the 300 negative samples. Then, demonstrations are randomly shuffled to eliminate positional effects, ensuring probes learn rule content rather than position patterns. We train simple layer-wise linear probes with L1-regularization \citep{alain2016understanding} on the residual streams of the LLMs. Specifically, for each layer $l$, we predict:

\begin{equation}
    \hat{y}^{(i)} = \sigma(\mathbf{w}^T \mathbf{h}_l^{(i)} + b)
\end{equation}

where $\mathbf{h}_l^{(i)}$ is the representation at layer $l$ for sample $i$, and we optimize over the regularized objective.

\subsection{Two-Scenario Evaluation Protocol}

We evaluate trained probes under two controlled scenarios to measure rule encoding dynamics: all Correct and only one corrupted. In the first scenario, all 4 demonstrations exhibit the same target rule (e.g., addition); in the second scenario, 3 demonstrations show the target rule, while 1 demonstration shows conflicting rule (e.g., multiplication or minus)

For each scenario, we extract representations from all layers and compute probe confidence as the measurement for model's encoded confidence of each rule. This yields four evaluation curves: Correct-rule probe confidence in all-correct scenario, correct-rule probe confidence in one-corrupted scenario, corrupted-rule probe confidence in all-correct scenario, and corrupted-rule probe confidence in one-corrupted scenario. To encourage the model to generate the encoded rule instead of the target answer directly, we modify the prompt to elicit direct operator prediction rather than numerical calculation. Specifically, for the Operator Induction tasks, we replace the query format from ``6 ? 3 = ?'' to:

\begin{verbatim}
What mathematical operation does ? represent? 
Choose from: plus, minus, multiplication

Answer:
\end{verbatim}

Figure \ref{fig:probe_analysis_detailed} illustrates these probe confidence patterns across three different model sizes. The results consistently demonstrate our key findings across model architectures. The probe analysis reveals two key mechanistic insights. First, in one-corrupted scenarios, both correct and corrupted rule probes exhibit above-baseline confidence, demonstrating that models encode multiple competing rules simultaneously rather than selecting a single rule early in processing.
Second, rule encoding emerges prominently in early-to-middle layers (layers 5-20), with confidence levels stabilizing in later layers. 

\section{Logit Lens Analysis Details}
\label{app:logit_lens}

\begin{figure}[t!]
    \centering
    \includegraphics[width=0.99\linewidth]{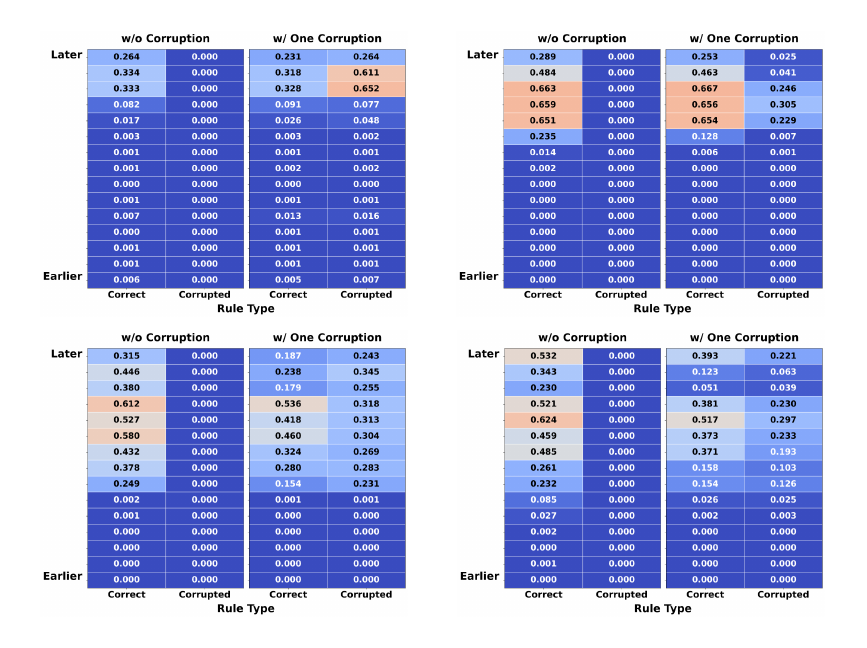}
    \caption{Logit lens results across model layers for Qwen3-0.6B, Qwen3-4B, Llama-3.1-8B-Instruct, and Llama-3.2-3B-Instruct from left to right, up to down.}
    \label{fig:lens_analysis_detailed}
\end{figure}

We employ logit lens analysis \citep{nostalgebraist2020logitlens} to track how models progressively resolve conflicting evidence during in-context learning. The logit lens projects intermediate representations from each transformer layer through the final layer normalization and language modeling head to observe predicted token probabilities at different processing stages.

For a hidden state $\mathbf{h}_\ell$ at layer $\ell$, the logit lens computes:
$$\mathbf{p}_\ell = \text{softmax}(\mathbf{W} \cdot \text{LayerNorm}(\mathbf{h}_\ell))$$

where $\mathbf{W}$ is the unembedding matrix of the language modeling head and $\mathbf{p}_\ell$ represents the probability distribution over vocabulary tokens as predicted from layer $\ell$.

Then, we focus specifically on correct or corrupted rules prediction by analyzing the probability assigned to the words of rules at each layer. Our analysis tracks two primary quantities as below:

\paragraph{Correct Rule Probability} The probability assigned to the word of ground-truth rule at layer $\ell$:
$$P_\ell(\text{correct}) = \max_{t \in T_{\text{correct}}} \mathbf{p}_\ell[t]$$
where $T_{\text{correct}}$ includes all tokens that match or prefix the word of correct rules \citep{wendler2024llamas}.

\paragraph{Corrupted Rule Probability} The probability assigned to the operator introduced through minority corruption:
$$P_\ell(\text{corrupted}) = \max_{t \in T_{\text{corrupted}}} \mathbf{p}_\ell[t]$$

We evaluate models under controlled corruption scenarios, exactly the same to the probes analysis, to measure conflict resolution dynamics internally. Wer show different models' logit lens decoding results in Figure \ref{fig:lens_analysis_detailed}. The logit lens analysis provides direct evidence for the second leg of our central hypothesis: models show strong signal of resolving conflicts in final layers.

\section{The Statistics of Two Types of Heads}

\begin{figure}[t]
    \centering
    \includegraphics[width=0.99\linewidth]{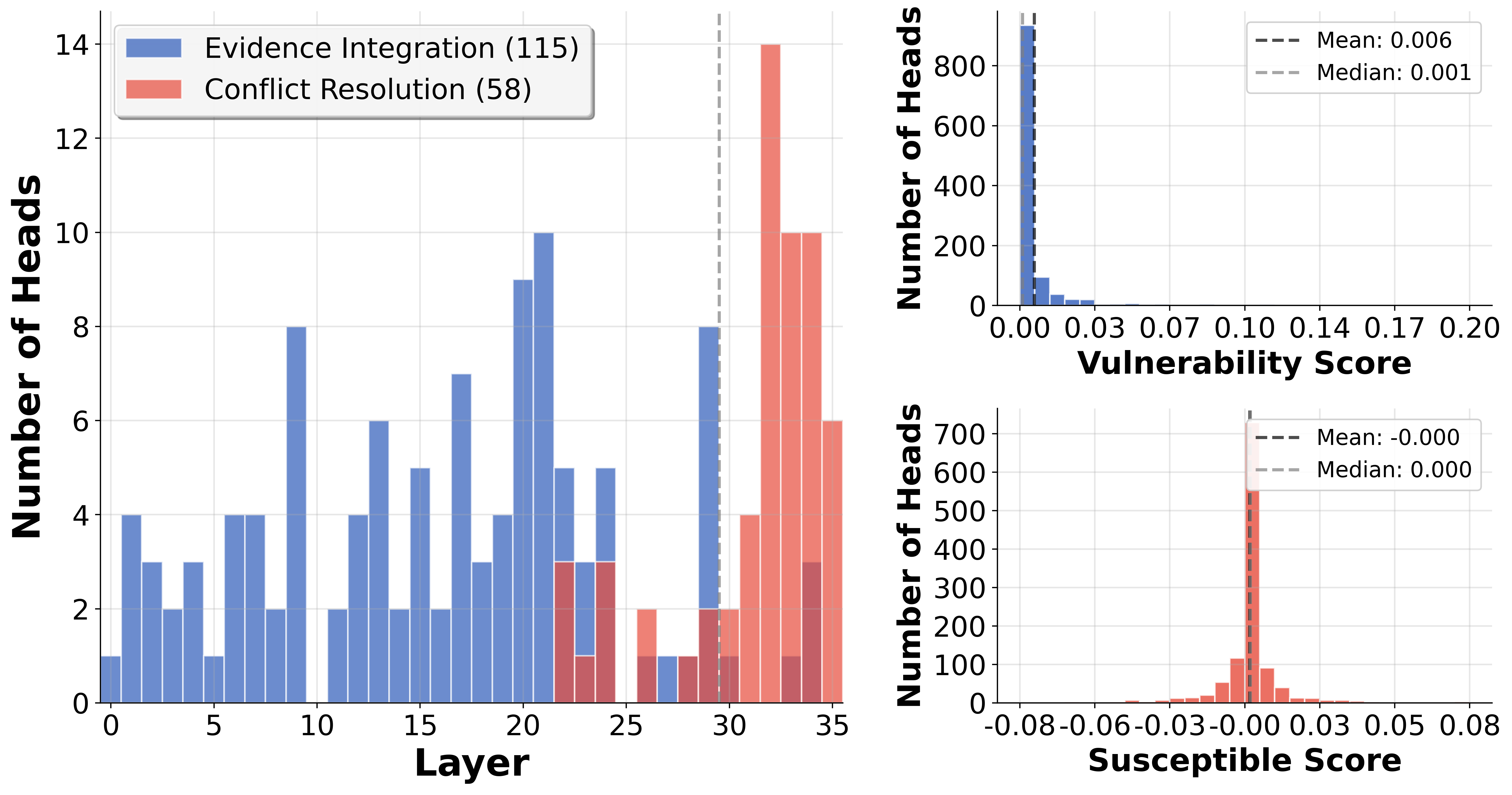}
    \caption{Distribution statistics of vulnerability and susceptible heads for Qwen3-4B model. }
    \label{fig:head_statistics}
\end{figure}

The layer-wise distribution in Figure \ref{fig:head_statistics} confirms the temporal separation predicted by our two-phase hypothesis, with vulnerability heads predominantly appearing in intermediate layers before 30 and susceptible heads concentrated in final layers. More importantly, the score distributions reveal that our metrics are highly selective rather than universal---the vast majority of attention heads exhibit near-zero scores (median values approach zero for both metrics), while only a small subset shows high vulnerability or susceptibility. This concentration of high scores among a minority of heads validates that our identification methodology successfully isolates functionally specialized components rather than capturing generic attention behaviors. Combined with our causal ablation results, these statistics provide strong evidence that the identified heads play specific, non-redundant roles in creating and failing to resolve conflicts during rule inference.

\section{Reasoning Uncertainty under Demonstration Conflicts}

Beyond classification accuracy, we also examine whether models exhibit calibrated reasoning uncertainty when facing demonstration conflicts. Since our rule inference tasks employ Chain-of-Thought (CoT) reasoning, we track next-token prediction entropy during CoT generation as a proxy for the model's internal uncertainty, following recent work showing that entropy reliably reflects and monitors reasoning confidence \citep{fu2025deep, kang2025scalable, zur2025language}. 

We measure the next-token entropy for Qwen3-4B on the Operator Induction queries under clean, corrupted, and corrupted-with-intervention conditions. As shown in Table \ref{tab:entropy_analysis} at Appendix \ref{app:additional}, demonstration conflicts significantly increase reasoning uncertainty: entropy rises from 0.0952 to 0.1161 nats, indicating the model are more obfuscated when facing conflicting information. Moreover, ablating identified heads partially restores confidence: masking the top-5 Vulnerability Heads and Susceptible Heads both reduce the next-token entropy, approaching the clean baseline. This demonstrates that our identified components causally contribute not only to prediction accuracy but also to elevated reasoning uncertainty, providing additional validation of their mechanistic roles through an orthogonal evaluation metric.

\section{Why Logits Attribution over Activation Patching for Locating Susceptible Heads}

Activation patching is another widely-used technique to identify causally important model components by targetedly replacing their activations and observing performance changes \citep{meng2022locating, zhang2023towards}. While our experimental setup naturally supports activation patching, where we have structured clean and corrupted demonstrations, we employ logits attribution to localize Susceptible Heads for the following reason. Activation patching fails to isolate conflict resolution mechanisms due to the temporal separation of conflict detection and conflict resolution functions. For example, patching heads that encode conflict with clean activations simply relieves the necessity of conflict resolution, consequently creating systematic false positives. In contrast, logits attribution respects this functional separation by measuring each head's direct contribution given the evidence it actually processes.


\section{Additional Results}

\label{app:additional}


\begin{wrapfigure}{r}{0.50\textwidth}
\centering
\includegraphics[width=0.49\textwidth]{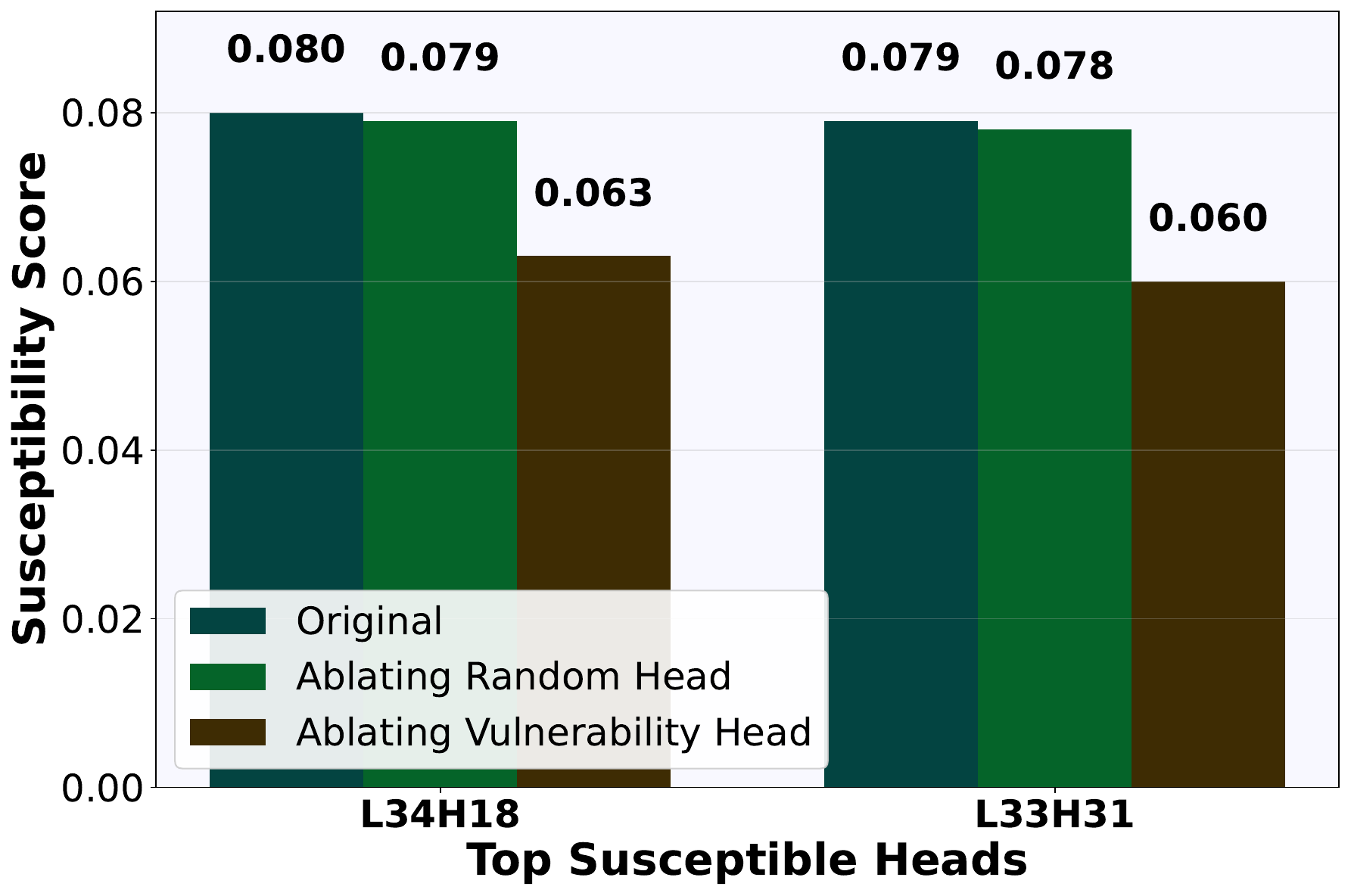}
\vspace{-0.2cm}
\caption{Ablating top vulnerability heads significantly reduces the susceptibility score of top susceptible heads.}
\label{fig:synergy}

\end{wrapfigure}

\begin{table}[t]
\centering
\caption{Average next-token prediction entropy during rule inference tasks under different conditions on Qwen3-4B. Lower entropy indicates higher model reasoning confidence.}
\vspace{0.2cm}
\label{tab:entropy_analysis}
\begin{tabular}{lcc}
\toprule
\textbf{Condition} & \textbf{Mean Entropy} & \textbf{Std} \\
\midrule
Clean & 0.0952 & 0.0239 \\
Corrupted & 0.1161 & 0.0217 \\
Corrupted + Vulnerability Heads masked & 0.1148 & 0.0167 \\
Corrupted + Susceptible Heads masked & 0.1120 & 0.0205 \\
\bottomrule
\end{tabular}
\end{table}

\begin{wraptable}{r}{0.50\textwidth}
\caption{Relative positional variance reduction (\%) when ablating vulnerability heads.}
\vspace{0.2cm}
\centering
\addtolength{\tabcolsep}{-2pt}
\begin{tabular}{cccc}
\hline
\multirow{2}{*}{\textbf{Models}} & \multicolumn{3}{c}{\textbf{\# of Ablated Heads}} \\
\cline{2-4}
& \textbf{5} & \textbf{8} & \textbf{10} \\
\hline
\textbf{Qwen3-0.6B} & 14.26 & 12.55 & 12.35 \\
\textbf{Qwen3-4B} & 18.45 & 22.50 & 19.73 \\
\textbf{Llama-3.1-8B} & 20.45 & 16.12 & 15.30 \\
\textbf{Llama-3.2-3B} & 13.50 & 16.80 & 11.95 \\
\hline
\end{tabular}
\label{tab:variance_reduction}
\end{wraptable}

\end{document}